\documentclass[1p]{elsarticle}

\usepackage{lineno,hyperref}
\usepackage{natbib}
\usepackage{geometry}
\usepackage{graphicx}
\usepackage{caption}
\usepackage{subcaption}
\usepackage[nohyperlinks,nolist]{acronym}
\usepackage[cmex10]{amsmath}
\usepackage{mathtools}
\usepackage{multicol}
\usepackage{todonotes}
\usepackage{multirow}
\usepackage{adjustbox}
\modulolinenumbers[5]
\usepackage{siunitx}
\journal{Reliability Engineering and System Safety}

\begin{acronym}
\acro{pdm}[PDM]{Predictive Maintenance}
\acro{rul}[RUL]{Remaining Useful Lifetime}
\acro{dnn}[DNN]{Deep Neural Network}
\acroplural{dnn}[DNNs]{Deep Neural Networks}
\acro{mmd}[MMD]{Maximum Mean Discrepancy}
\acro{cnn}[CNN]{Convolutional Neural Network}
\acroplural{cnn}[CNNs]{Convolutional Neural Networks}
\acro{rnn}[RNN]{Recurrent Neural Network}
\acro{lstm}[LSTM]{Long Short Term Memory Network}
\acro{mlp}[MLP]{Multi-Layer Perceptron}
\acro{da}[DA]{Domain Adaption}
\acro{cmapss}[C-MAPSS]{Commercial Modular Aero-Propulsion System Simulation}
\acro{rmse}[RMSE]{Root Mean Squared Error}
\acro{mse}[MSE]{Mean Squared Error}
\acro{dann}[DANN]{Domain Adversarial Neural Network}
\acro{hpc}[HPC]{high-pressure compressor}
\acro{umap}[UMAP]{Uniform Manifold Approximation and Projection}
\acro{grl}[GRL]{Gradient Reversal Layer}
\acro{relu}[ReLU]{Rectified Linear Unit}
\acro{ssl}[SSL]{Semi-Supervised Learning}
\acro{ae}[AE]{Autoencoder}
\acro{rbm}[RBM]{Restricted Boltzmann Machine}
\acro{iqr}[IQR]{interquartile range}
\end{acronym}

\begin{document}

\begin{frontmatter}

\title{Improving Semi-Supervised Learning for Remaining Useful Lifetime Estimation Through Self-Supervision}

\author[1]{Tilman Krokotsch\corref{cor1}}
\ead{tilman.krokotsch@tu-berlin.de}

\author[2]{Mirko Knaak}

\author[1]{Clemens Gühmann}

\cortext[cor1]{Corresponding author}

\address[1]{Chair of Electronic Measurement and Diagnostic Technology, 
            Technische Universität Berlin}
\address[2]{Thermodynamics \& Power Systems, 
            Power Train \& Power Engineering, 
            IAV GmbH}

\begin{abstract}
RUL estimation suffers from a severe data imbalance where data from machines near their end of life is rare.
Additionally, the data produced by a machine can only be labeled after the machine failed.
\ac{ssl} can incorporate the unlabeled data produced by machines that did not yet fail.
Previous work on \ac{ssl} evaluated their approaches under unrealistic conditions where the data near failure was still available.
Even so, only moderate improvements were made.
This paper proposes a novel \ac{ssl} approach based on self-supervised pre-training.
The method can outperform two competing approaches from the literature and a supervised baseline under realistic conditions on the NASA \acl{cmapss} dataset.
\end{abstract}


\end{frontmatter}

\section{Introduction}
\label{sec:Introduction}
    \ac{pdm} is one of the core pillars of Industry 4.0 and enables more cost-effective operation of machinery.
    While early approaches to \ac{pdm} focused on hand-crafted, physical models and heuristics, nowadays data-driven methods are on the rise.
    Fueled by massive amounts of data provided by an increasing number of sensors, data-driven \ac{pdm} makes hand-crafting physical models less necessary.
    Due to the advent of deep learning, data-driven models can ingest even more data without the need for specialized feature engineering.
    Nevertheless, \ac{pdm} suffers from a server data imbalance as data from healthy machines is far more ubiquitous than data from degraded or faulty ones.
    This makes training effective data-driven models a challenging task.

    \ac{rul} estimation, as a sub-field of \ac{pdm}, is defined as “the length from the current time to the end of the useful life”. \cite{Si2011}
    It plays a vital role in effectively scheduling maintenance operations.
    Unfortunately, labeling data for \ac{rul} estimation is only possible after a machine fails, making it hard to acquire enough labeled data for conventional, supervised approaches to work.
    On the other hand, large amounts of unlabeled data are available from machines that did not yet fail.
    \ac{ssl} can be a possible solution for this problem.

    \acf{ssl} uses labeled and unlabeled data in concert to solve a learning problem \cite{VanEngelen2020}.
    It makes it possible to distill knowledge from large amounts of unlabeled data, thus lowering the amount of labeled data needed to achieve good performance.
    Specifically, \ac{ssl} aims to learn a conditional distribution $P(y|x)$ where $x\sim P(X)$ are the available features (i.e. sensor readings) and $y\sim P(Y)$ are the labels (i.e. the \ac{rul}).
    The learning algorithm has access to the set of labeled training data $D_L = \{(x_1, y_1), \dots, (x_i, y_i)\}$ and the mutually exclusive set of unlabeled data $D_U=\{x_{i+1}, \dots, x_{i+j}\}$.
    
    Recent works, e.g. \cite{ListouEllefsen2019,Yoon2017}, have shown promising results using different \ac{ssl} methods for \ac{rul} estimation.
    More specifically, they were able to modestly reduce the test \ac{rmse} and \ac{rul}-Score on subsets of the NASA \ac{cmapss} dataset using as few as 1\% of the labeled data.
    Although these findings are leading in the right direction, there are some shortcomings this paper wants to address.
    First, the previous work only evaluates their \ac{ssl} approaches on one of the four subsets of the \ac{cmapss} dataset.
    As these subsets are relatively small (compared to other deep learning data sets), at least all of the subsets should be investigated to see if the performance improvements are universal.
    Secondly, the time series of the unlabeled data were used as-is.
    In our previous work \cite{Krokotsch2020}, we have shown that unlabeled time series data for \ac{rul} estimation should not contain time steps at the point of failure.
    If they do, the whole time series could be labeled trivially, making the use of \ac{ssl} unnecessary.
    Therefore, the previous studies on \ac{ssl} for \ac{rul} estimation may produce overly optimistic results, as the approaches have access to more data near failure than could be considered realistically.
    This \emph{grade of degradation} of the unlabeled data, i.e. how far from failure the machines in it are, is of significant importance for \ac{ssl}.
    
    As previous work produced only modest improvements, we propose a novel \ac{ssl} method based on self-supervised pre-training.
    Self-supervision has been proven useful for pre-training deep neural networks on unlabeled data \cite{Doersch_2015_ICCV,gidaris2018unsupervised,Devlin2019}.
    Our pre-training task is to estimate the time between two time steps in a time series as a proxy for the difference in \ac{rul}.
    Afterward, the pre-trained network is fine-tuned on labeled data.
    We will conduct experiments on all four subsets of \ac{cmapss} comparing our approach and two common \ac{ssl} approaches, i.e. \acp{rbm} and \acp{ae}, to a supervised baseline.
    This will serve to answer our following research questions:
    
    \begin{itemize}
        \item Do findings of previous \ac{ssl} studies on \acs{rul} estimation hold when taking the \emph{grade of degradation} of the unlabeled data into account?
        \item Can self-supervised pre-training improve on other pre-training-based \ac{ssl} approaches for \acs{rul} estimation?
    \end{itemize}
    
    The remaining paper is structured as follows.
    First, we will lay out the related work on \ac{rul} estimation with deep neural networks, \acl{ssl} and self-supervised learning in section \ref{sec:RelatedWork}.
    In section \ref{sec:Methods}, we will describe our network architecture, our semi-supervised learning approach, and our self-supervised pre-training.
    Afterward, we explain our experimental setup, including the data, performance metrics, and competing approaches, as well as the evaluation procedure and hyperparameter selection.
    Section \ref{sec:Results} is concerned with the presentation and discussion of our results, while section \ref{sec:Conclusion} concludes this paper and gives an outlook on future work.
    The code to reproduce the results of this paper is available on GitHub: \texttt{\url{www.github.com/tilman151/self-supervised-ssl}}.

\section{Related Work}
\label{sec:RelatedWork}
    This section gives an overview of the current state of the literature.
    First, we will discuss \ac{rul} estimation with \acp{dnn}.
    Second, we will investigate \acl{ssl} and self-supervised learning in the scope of \ac{rul} estimation.

    \subsection{Remaining Useful Lifetime Estimation}
    \label{sec:RW:RULEstimation}
        \ac{rul} estimation is often treated as a regression problem.
        Recent work focuses mainly on \acp{dnn} because they work on the raw data and do not require hand-crafted features.
        Obviously, the early works were focuses on shallow networks and \acp{mlp} \cite{Gebraeel2004}.
        Later ones settled on \acp{cnn} and \acs{lstm} as network architectures.
        
        \acp{lstm} are \acp{rnn} and a natural fit for the time series data seen in \ac{rul} applications.
        They process one time step at a time with the help of an internal cell state derived from all previously seen time steps.
        Zheng et al. \cite{Zheng2017} used \acp{lstm} on three benchmark datasets and found them working best compared to \acp{mlp} and \acp{cnn}.
        Wu et al. \cite{Wu2018} compared \acp{lstm} against vanilla \acp{rnn} and GRUs.
        They declared \acp{lstm} the winner, too.
        
        \acp{cnn} on the other hand seem better suited for image data than time series.
        But, using 1d-convolution instead of 2d we can use it for \ac{rul} estimation, too.
        In a comparison, Bai et al. \cite{Bai2018} found \acp{cnn} equal to \acp{lstm} in performance, even though they are faster in inference and training.
        First attempts at \ac{rul} estimation from Babu et al. \cite{SateeshBabu2016} were still worse than \acp{lstm}.
        They were still using 2d-convolution and when Li et al. \cite{Li2018a} switched to 1d, they were able to surpass \acp{lstm} altogether.
        Zhu et al. \cite{Zhu2019} combined features from multiple hidden layers with their multiscale \ac{cnn} which did better on their dataset than traditional \acp{cnn}.
        Instead of using raw time series data, they transformed it to the frequency domain and used that as the input of their network.
        Jiang et al. \cite{Jiang2020} resorted to combining both network types and report better performance.
        
        Next to all networks were trained against \ac{mse} \cite{Zheng2017,Wu2018,SateeshBabu2016,Zhu2019,Jiang2020}.
        Only Li et al. \cite{Li2018a} used \ac{rmse}.
    
    \subsection{Semi-Supervised Learning}
    \label{sec:RW:SSL}
        \ac{ssl} can be divided into several sub-fields \cite{VanEngelen2020}.
        The most common distinction is the goal of the training process itself.
        While transductive methods are only concerned with providing the labels for the unlabeled training data, inductive methods yield a model that can be used on unseen data.
        We will focus on the inductive methods, as for \ac{pdm} applications we need to apply the trained model on unseen data after training.
        Even though there are many diverse approaches to \ac{ssl} (e.g. S3VMs), we will narrow our perspective to the ones applicable to \acp{dnn}.
        
        Wrapper methods offer some of the oldest \ac{ssl} approaches.
        They rely on producing pseudo-labels for the unlabeled portion of the data and then using it in conjunction with the labeled data to train supervised.
        Self-training is one of the most basic methods and was published by Yarowsky et al. in 1995 \cite{Yarowsky1995}.
        First, a model trained only on the labeled data provides the pseudo-labels for the unlabeled data.
        Afterward, a final model is trained on the combined labeled and pseudo-labeled data.
        
        Pre-training-based methods rely on unsupervised learning.
        The \ac{dnn} or a part of it is trained with an unsupervised learning method and taken as the initialization for a supervised learning stage.
        The literature provides examples for several unsupervised approaches, most commonly \acp{ae}, used for pre-training.
        Cheng et al. \cite{Cheng2019} used deep autoencoders for semi-supervised machine translation.
        Kingma et al. \cite{Kingma2014semi} used variational autoencoders for semi-supervised image classification.
        
        There are several deep learning methods for \ac{ssl} that directly incorporate an unsupervised component into their loss.
        Examples are Ladder Networks \cite{Rasmus2015semi} which incorporate an autoencoder reconstruction loss, or Pseudo-ensembles \cite{Bachman2014semi} which use a consistency loss between the outputs of a parent network and perturbed child networks for the unlabeled data. The survey of Van Engelen et al. \cite{VanEngelen2020} gives an excellent overview of the previously mentioned methods.
        
        There are a few papers on \ac{ssl} for \ac{rul} estimation.
        Ellefsen et al. \cite{ListouEllefsen2019} used a \ac{rbm} for pre-training on the NASA \ac{cmapss} dataset.
        Yoon et al. \cite{Yoon2017} pre-trained their network on the same dataset with a variational autoencoder.
        He et al. \cite{He2018rul} used ladder networks for \ac{rul} estimation of centrifugal pumps.
        Unfortunately, the field of \ac{ssl} suffers from a multitude of evaluation setups, which makes comparing approaches difficult \cite{Oliver2018realistic}.
        \ac{ssl} for \ac{rul} estimation is not excluded from this issue, as many papers report results only on parts of their datasets, or omit critical information like the amount of available labeled data.
        We aim to take these pitfalls into account in this work.
    
    \subsection{Self-Supervised Learning}
    \label{sec:RW:SelfSupervised}
        Like many methods in deep learning, self-supervised learning first succeeded in computer vision and spread to different fields from there.
        The aim was to learn features that are beneficial for solving common tasks (image classification, object detection, etc.) in the absence of labeled data.
        For this, a so-called pre-text task is defined to train a neural network, which is then used as a feature extractor for the real task.
        For example, Doersch et al. \cite{Doersch_2015_ICCV} predicted the position of an image patch, cut from a larger image, in relation to another patch from the same image.
        The trained network was then able to perform object detection.
        Another example is Gidaris et al. \cite{gidaris2018unsupervised}, who used predicting image rotation as a pre-text task for classification.
        Approaches like these were able to produce state-of-the-art results in a semi-supervised regime (large amount of unlabeled, small amount of labeled data) but could not outperform supervised approaches trained on large-scale datasets.
        
        Self-supervised learning gained prominence as a pre-training technique in natural language processing through Delvin et al. \cite{Devlin2019}.
        Their model, named BERT, was pre-trained on the pre-text task of predicting missing words in sentences sampled from a $3.3$M word dataset, including the whole of the English Wikipedia.
        The model was then able to produce state-of-the-art results on eleven benchmark tasks by training a single layer on top of the pre-trained network.
        
        Metric learning is another possible pre-text task that can be used supervised and self-supervised.
        It aims to learn a distance or similarity metric between pairs of data points.
        The recent work of Musgrave et al. \cite{Musgrave2020} gives an excellent overview of popular methods but shows that newer, more complex approaches perform only as good as older, simple ones.
        This leads us to the conclusion that even simple metric learning methods could be used as a self-supervised pre-text task, too.
        
        There is, to our knowledge, still a lack of work in self-supervised methods for multivariate time series data as found in \ac{pdm} applications.
        Recently, Franceschi et al. \cite{Franceschi2019} proposed a metric-learning-based pre-text task for time series.
        They trained siamese networks \cite{baldi1993neural,bromley1993signature} to predict the similarity of two time series snippets with a triplet loss.
        Their pre-trained network outperformed dynamic time warping in an unsupervised regime, other deep learning approaches in a semi-supervised regime and yield competitive results when compared to supervised state-of-the-art approaches.
        This hints at self-supervised learning as a promising direction for pre-training on time series data.

\section{Methods}
\label{sec:Methods}
    In this section, we will describe the neural network we use for \ac{rul} estimation, \ac{ssl} via pre-training, and our novel self-supervised pre-training technique.

    \begin{figure}
        \centering
        \begin{subfigure}[b]{0.49\textwidth}
            \includegraphics[width=\textwidth]{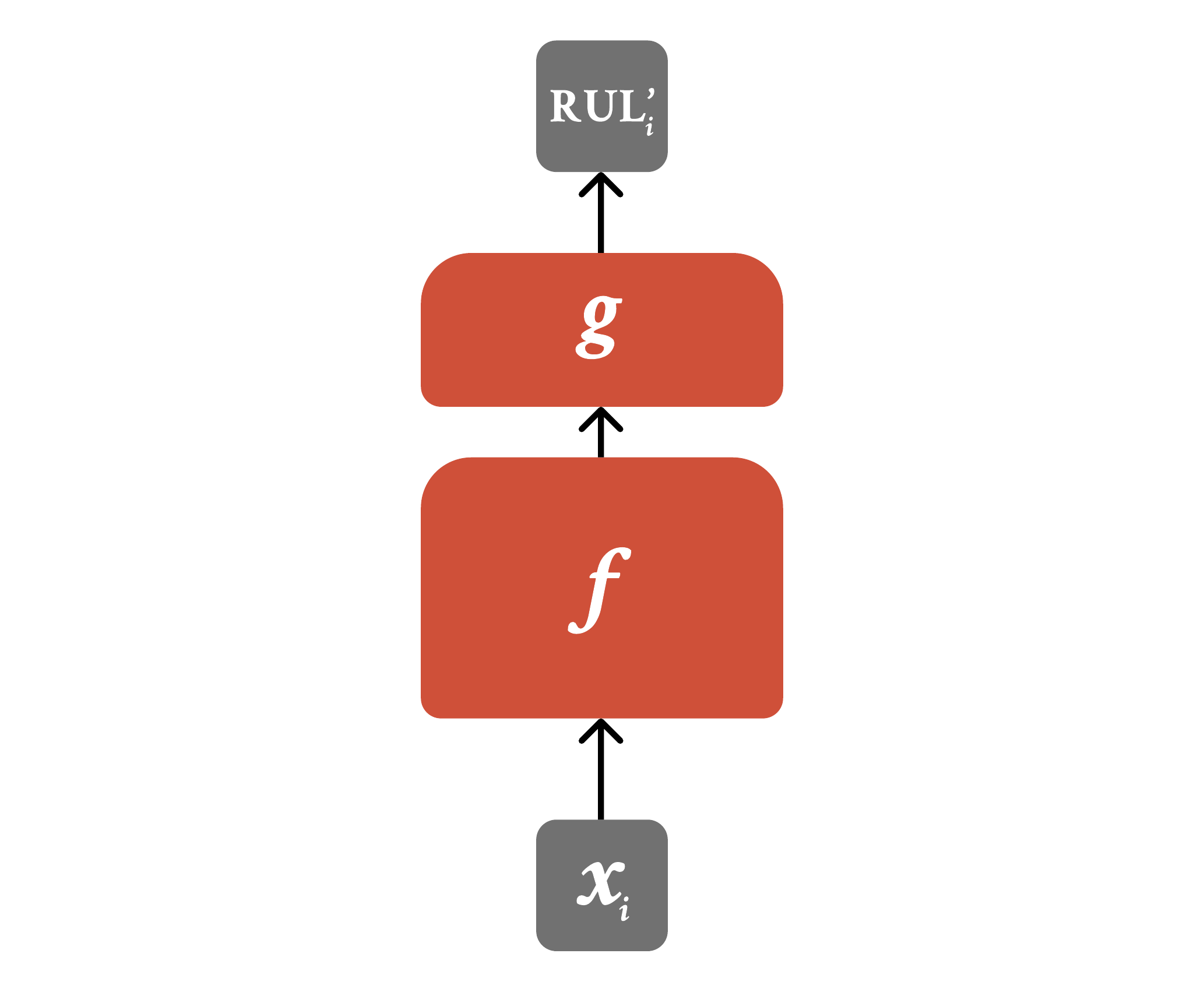}
            \caption{\ac{rul} Estimation Network}
            \label{fig:RulEst}
        \end{subfigure}
        \begin{subfigure}[b]{0.49\textwidth}
            \includegraphics[width=\textwidth]{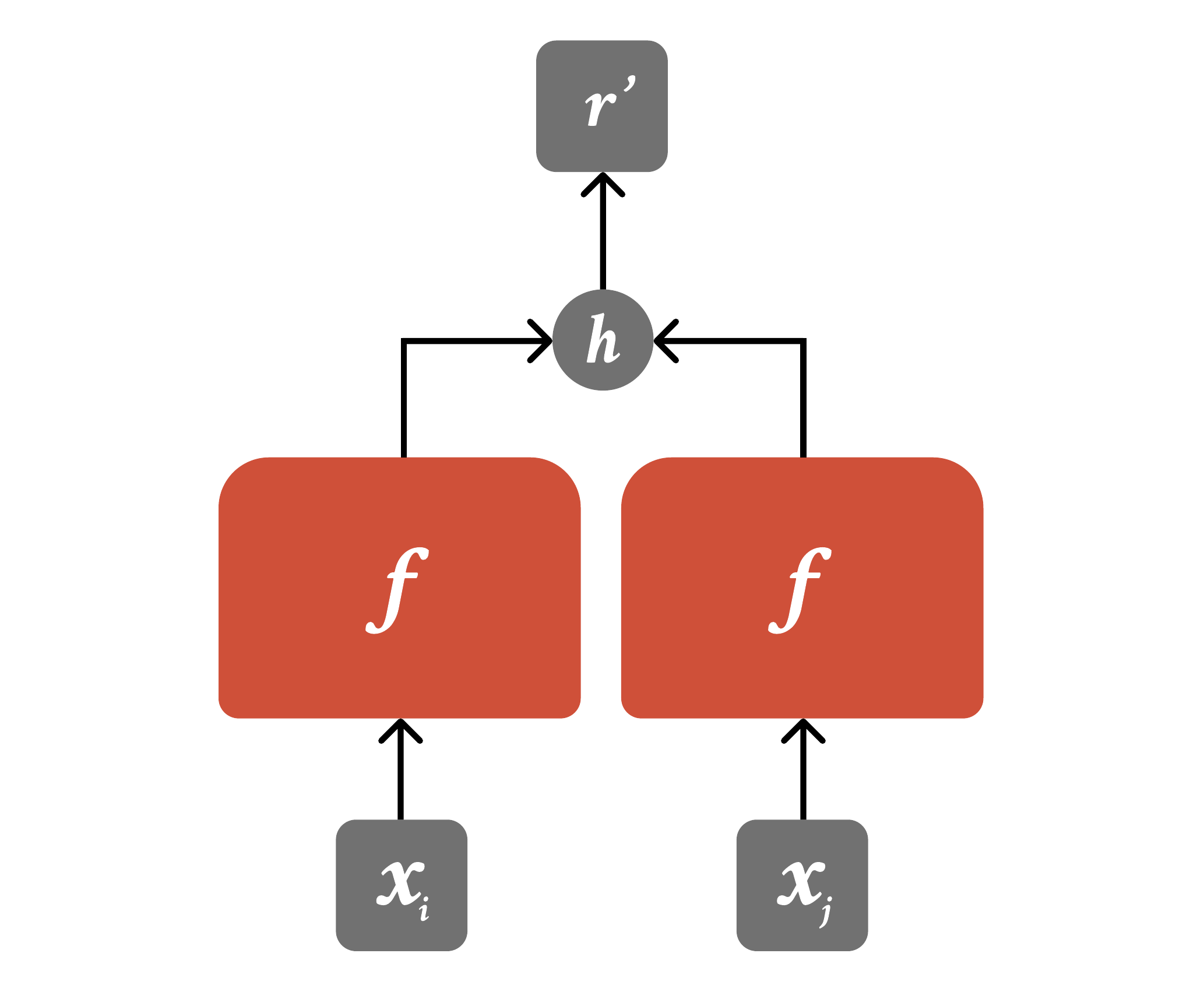}
            \caption{Self-Supervised Siamese Network}
            \label{fig:PreTrain}
        \end{subfigure}
        \caption{
            \textbf{Overview of the networks used in this work:} 
            Depicted are the feature extractor $f$, the regression head $g$, and the distance function $h$. 
            The supervised \ac{rul} estimation network takes a time frame $x_i$ as its input and predicts the \ac{rul} value $\textsc{RUL}_i'$. 
            The siamese network produces a pair of embeddings from the inputs $x_i$ and $x_j$. 
            The distance $h(f(x_i), f(x_j))$ is $r'$, the predicted relative \ac{rul}. 
            The trainable parts of the networks are depicted in red.
            }
        \label{fig:Networks}
    \end{figure}

    \subsection{Remaining Useful Lifetime Estimation Network}
    \label{sec:M:RULEstimation}
    
        \begin{figure}
            \centering
            \includegraphics[width=\textwidth]{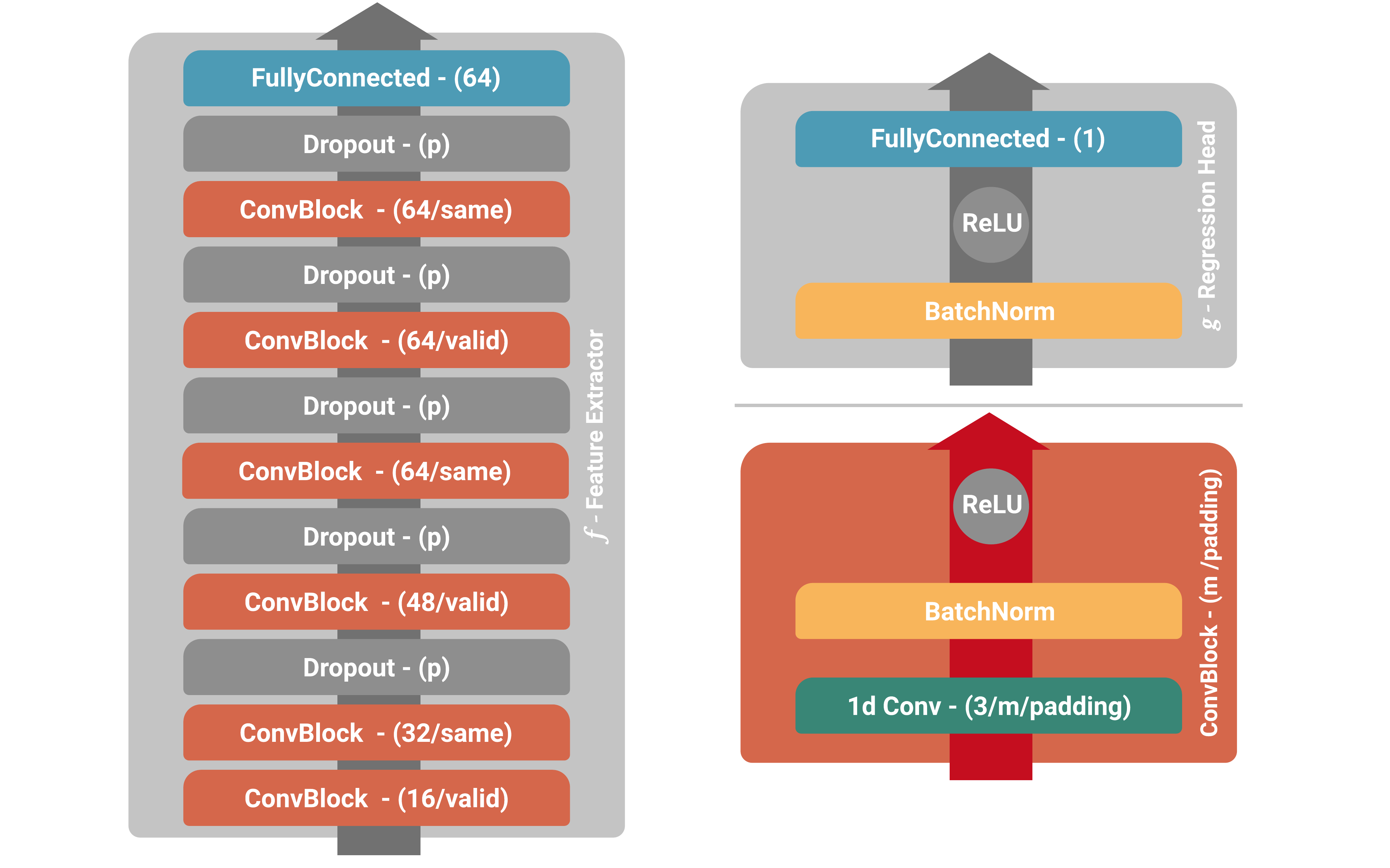}
            \caption{\textbf{Architecture of the used Networks:} The feature extractor $f$ consists of six $ConvBlocks$. These blocks are stacks of 1d-convolutions of kernel size three, $m$ filters, and either $same$ or $valid$ convolution padding with zeros, followed by Batch Normalization and a \ac{relu} activation function. The even layers use valid padding and the odd layers use same padding. This makes it possible to use more $ConvBlocks$ with valid padding only as same padding does not reduce the size of the input. In between the $ConvBlocks$, we place 2d-dropout that drop whole time steps from their input. The probability of dropping a time step is $p$. The last layer of the feature extractor is a fully-connected layer reducing its input to a latent dimension of 64. The regression head $g$ is a Batch Normalization layer followed by a \ac{relu} activation function and a fully-connected regression layer with a single unit.}
            \label{fig:BasicArchitecture}
        \end{figure}
    
        In general, \acp{dnn} for regression follow a common architecture, as seen figure \ref{fig:RulEst}.
        They consist of two networks: the feature extractor $f$ and a regression head $g$.
        Networks for \ac{rul} estimation are no different, estimating the \ac{rul} value of a sample $x$ as:
        \begin{equation}
            \textsc{RUL}' = g(f(x))
        \end{equation}
        As seen in section \ref{sec:RW:RULEstimation}, the feature extractor can take the form of a \ac{cnn}, \ac{lstm} or any other network architecture.
        The regression head, on the other hand, almost always takes the form of a simple linear layer ($ax+b$) or a \ac{mlp}.
        
        We expect the network input $x_{(i-w):i}^{(k)}$, or $x_i$ for short, to be a time frame of size $w$ from a multivariate time series $k$ ending at time step $i$.
        Therefore, we use a simple 1d-\ac{cnn} feature extractor with a fully-connected layer as the regression head.
        The feature extractor consists of multiple 1d-\ac{cnn} layers with Dropout, Batch Normalization \cite{Ioffe15}, and a \ac{relu} activation function, followed by a single linear layer also with Batch Normalization and \ac{relu}.
        The complete architecture is depicted in figure \ref{fig:BasicArchitecture}.
        We selected a \ac{cnn} because they are faster to train than e.g. \acp{lstm} and have less trainable parameters than a \ac{mlp} with similar performance.
        In this work, we will focus on this feature extractor only, as we are mainly concerned with the influence of pre-training it on unlabeled data.
        Furthermore, previous works imply that differences in performance between different extractor architectures for \ac{rul} estimation are marginal \cite{Li2018a}.
        
        We train the network by mini-batch gradient descent with a \ac{rmse} loss:
        \begin{equation}
            \mathcal{L}_{\textsc{RMSE}} = \sqrt{\frac{1}{|X|}\sum_{i=1}^{|X|} (\textsc{RUL}_i - \textsc{RUL}'_i)^2}
            \label{eqn:rmse}
        \end{equation}
        where $X$ is a (mini-)batch of samples $\{x_1, ..., x_{|X|}\}$, $\textsc{RUL}_i$ the \ac{rul} value associated with $x_i$ and $\textsc{RUL}'_i = g(f(x_i))$.

    \subsection{Semi-Supervised Learning through Pre-Training}
    \label{sec:M:SSL}
        \ac{ssl} is a machine learning regime that includes labeled and unlabeled data into the training process.
        In this paper, we focus on the two-stage approach of combining unsupervised pre-training with supervised fine-tuning.
        First, the feature extractor $f$ is trained using a method that does not require labels on the features of the labeled and unlabeled data.
        Afterward, the pre-trained feature extractor is used as the starting point for the conventional, supervised training as described in the previous section.
        The regression head $g$ is still initialized randomly.
        
        This procedure aims for the feature extractor to learn useful features from the unsupervised data that are beneficial to the supervised training.
        For this to work, it is necessary to assume that the marginal data distribution $P(X)$ contains information about the conditional distribution $P(\textsc{RUL}|X)$ we want to learn \cite{VanEngelen2020}.
        It follows that we have to assume that the labeled and unlabeled data were generated by the same distribution $P$.

    \subsection{Self-Supervised Pre-Training}
    \label{sec:M:SelfSupervised}
        In this section, we lay out the intuitive motivation behind our proposed self-supervised pre-training through analyzing the latent space of trained networks.
        Afterward we derive a pre-text task from this intuition and devise a training regime.

        \subsubsection{Motivation}
        \label{sec:M:SS:Motivation}
            To be able to design a suitable pre-text task, one has to understand the supervised training process and its shortcomings.
            To gain an understanding of the supervised training process, we will look at the features learned by the feature extractor $f$ trained with different amounts of labeled data.
            Unfortunately, high-dimensional data, such as the extracted features, is impossible to understand for humans.
            Therefore, we project it into two dimensions by using \ac{umap} \cite{mcinnes2018umap}, and plot the features in a scatter plot.
            
\begin{figure}
                \begin{subfigure}[b]{0.49\textwidth}
                \centering
                \includegraphics[width=\textwidth]{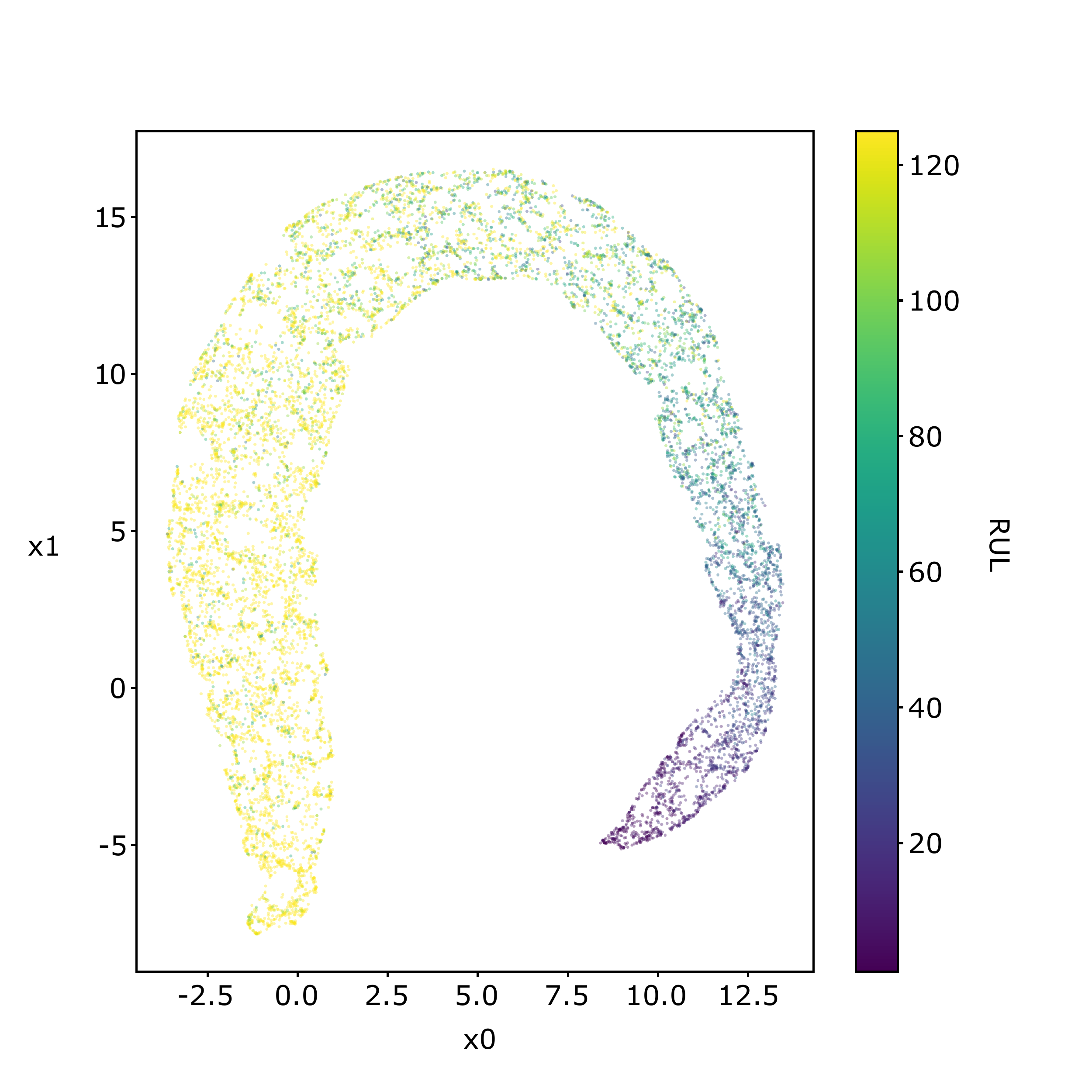}
                \caption{All labeled time series}
                \label{fig:embeddings_full}
                \end{subfigure}
                \begin{subfigure}[b]{0.49\textwidth}
                \centering
                \includegraphics[width=\textwidth]{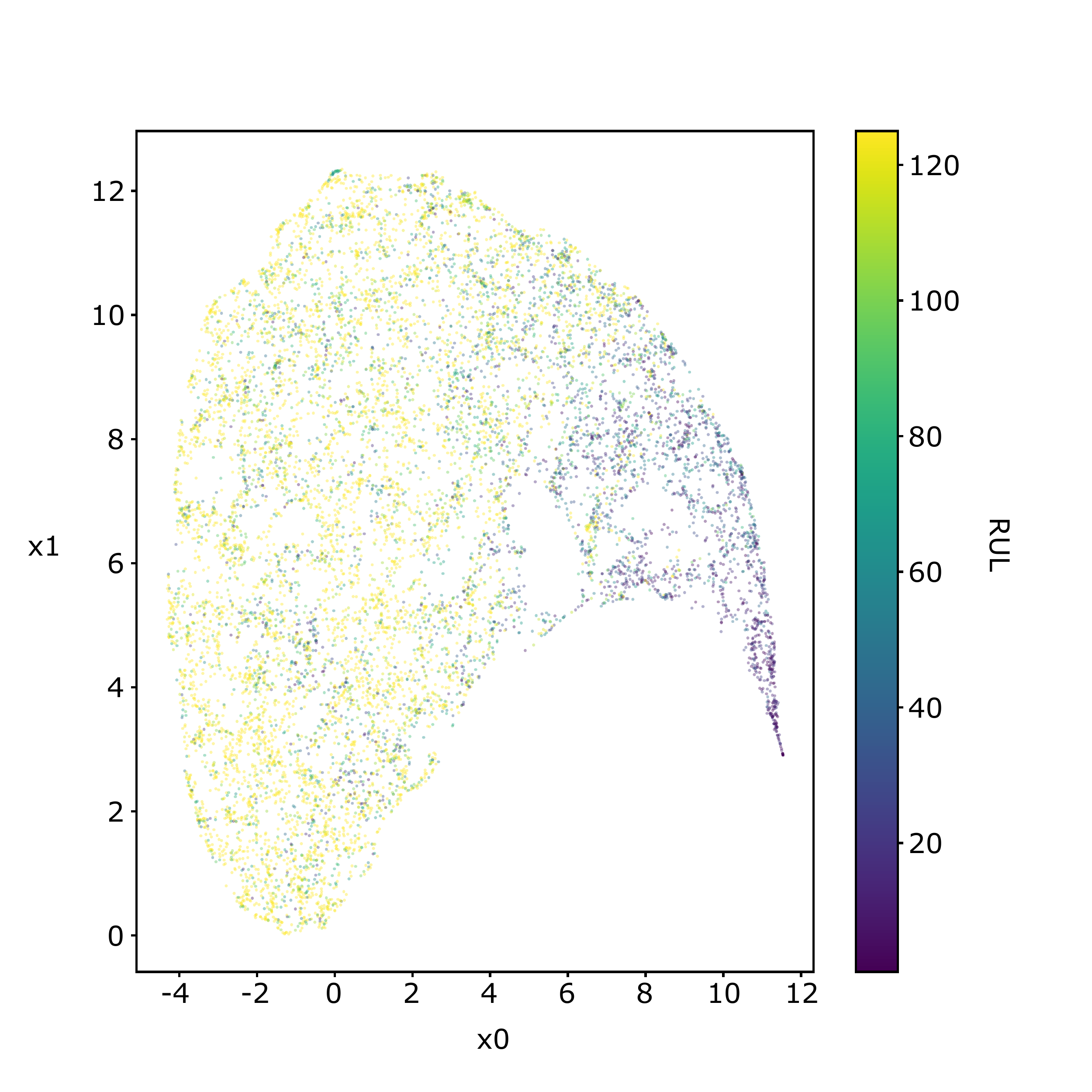}
                \caption{Three labeled time series}
                \label{fig:embeddings_few}
                \end{subfigure}
                \caption{\textbf{Features produces by trained feature extractors:} The depicted features are \ac{umap} \cite{mcinnes2018umap} projections of the output of a feature extractor $f$. The first plot shows an $f$ trained on all labeled data from FD004, while the second one shows an $f$ trained on only three time series. Both were trained for 10 epochs. The color indicates the \ac{rul} value associated with the embedding. Both (a) and (b) were generated from the validation data of FD004. The random seed for both training runs was fixed.}
                \label{fig:embeddings}
            \end{figure}

            Figure \ref{fig:embeddings_full} depicts the features extracted from the validation data of \ac{cmapss} subset FD004 after training on all available labeled  data.
            The network achieves a validation \ac{mse} loss of $19.3$.
            We can observe a snake-like structure in the features that broadens on the left side and narrows on the right side.
            Though coloring the data points according to their associated \ac{rul} value, we can see that the left end of the snake corresponds to high and the right side to low \ac{rul} values.
            Furthermore, no discernible sub-structures are apparent which shows that the network does not differentiate between time frames from different time series and the same \ac{rul}.
            Because of the low validation loss, we can conclude that this snake-like structure in the feature space may be beneficial to \ac{rul} estimation.
            
            Figure \ref{fig:embeddings_few} depicts the features extracted by a feature extractor trained only on three labeled time series, which corresponds to $2\%$ of the available labeled data.
            The network achieves a validation loss of $31.8$.
            Again, we can observe the snake-like structure but this time the high-\ac{rul} end is much more feathered out.
            Additionally, the \ac{rul} values are not as clearly separated as in the previous figure.
            
            Comparing the two plots, we can theorize that the network fails in the low data scenario because it cannot learn the tight snake-like structure.
            A suitable pre-text task should result in a feature extractor that already learned this structure which we can verify qualitatively with a plot like in figure \ref{fig:embeddings}.
            Data points with similar \ac{rul} values should be near each other in the feature space.
            
        \subsubsection{Definition of the Pre-Text Task}
        \label{sec:M:SS:Loss}
            To derive a pre-text task from the intuition gained in the previous chapter, we turn to the field of metric learning.
            Metric learning is concerned with learning a distance or similarity metric between two data points of a dataset.
            As we are aiming to learn self-supervised, we can use only information from the unlabeled data to achieve this task.
            In our case, we want to learn a latent space structure, where data points with similar \ac{rul} values are close and points with a large difference in \ac{rul} are distant from each other.
            This corresponds to predicting the \emph{relative \ac{rul}} between two data points.
            Given two time series, $k$ and $l$, of data from machines run to failure, the \emph{relative \ac{rul}} value $r$ is:
            \begin{equation}
                r = \textsc{RUL}^{(k)}_i - \textsc{RUL}^{(l)}_j
                \label{eqn:relRul}
            \end{equation}
            where $\textsc{RUL}^{(k)}_i$ is the \ac{rul} at time step $i$ in time series $k$.
            If we declare $k = l$ and given the assumption of a linear degradation model of:
            \begin{equation}
                \textsc{RUL}^{(k)}_i = |k| - i
            \end{equation}
            where $|k|$ is the length of time series $k$, we can rewrite equation \ref{eqn:relRul} as:
            \begin{equation}
                r = (|k| - i) - (|k| - j)
            \end{equation}
            It is trivial to see that $|k|$ is not needed at all to calculate $r$:
            \begin{eqnarray}
                r &=& j - i
            \end{eqnarray}
            This makes it possible to calculate $r$ even on time series of unfailed machines where $|k|$ is not known.
            Therefore, this pre-training target can be used without any failure data and consequently without the true \ac{rul} values.
            
            For a piece-wise linear degradation model as seen in the \ac{cmapss} dataset:
            \begin{equation}
                \textsc{RUL}^{(k)}_i = \max{(\textsc{RUL}_\textsc{max},\; |k| - i)}
            \end{equation}
            the formula for $r$ is still an acceptable approximation, if we assume that $i < j$ and $i - j \leq \textsc{RUL}_\textsc{max},\; \forall i,j$.
            We normalize $r$ to the range of $[0,1]$ by dividing it by $\textsc{RUL}_\textsc{max}$.
            The final equation is then:
            \begin{equation}
                r \approx \frac{j - i}{\textsc{RUL}_\textsc{max}},\quad \begin{aligned}i &< j \\ i - j &\leq \textsc{RUL}_\textsc{max} \end{aligned}\quad \forall i,j
                \label{eqn:finalTarget}
            \end{equation}

        \subsubsection{Training Procedure}
        \label{sec:M:SS:Training}
            The goal of our pre-training is training a neural network on estimating the target value $r$ from equation \ref{eqn:finalTarget} from which we then extract the feature extractor $f$ to be used to initialize the \ac{rul} estimation network before supervised training.
            We realize this goal with siamese networks \cite{bromley1993signature,baldi1993neural}.
            Siamese networks take the form of the feature extractor $f(x)$ and a distance function $h(a, b)$ operating on two samples, $x_i$ and $x_j$, so that:
            \begin{equation}
                r' = h(f(x_i), f(x_j))
            \end{equation}
            where $r'$ is the predicted value of $r$.
            The function $h(a, b)$ is defined as:
            \begin{equation}
                h(a, b) = \left|\left|\frac{a}{||a||} - \frac{b}{||b||}\right|\right|^2
            \end{equation}
            where $||\cdot||$ is the euclidean norm.
            Dividing the embeddings by their norm restricts the embedding space to a hyper ball, which often is found to be useful \cite{Musgrave2020}.
            Figure \ref{fig:PreTrain} depicts the schematic structure of the siamese networks.
            We use \ac{mse} as a loss function with mini-batch gradient descent to train the siamese networks:
            \begin{equation}
                \mathcal{L}_{\textsc{MSE}} = \frac{1}{|X|} \sum_{k=1}^{|X|}{(r_k - r_k')}
            \end{equation}
            For each (mini-)batch $X$, we sample $|X|$ time series from the training data, uniformly sample pairs of time frames $x^{(k)}_i$ and $x^{(k)}_j$, and calculate $r_k$ according to equation \ref{eqn:finalTarget}.
            Because the samples $x^{(k}_i$ and $x^{(k}_j$ are time frames from the same time series, a difference between $i$ and $j$ that is much smaller than the length of the time frame results in significant overlap between the samples.
            Therefore, we introduce a minimum distance for sampling pairs, which is regarded as a hyperparameter.
            We used the Adam optimizer \cite{Kingma2014a} for training with the default values of $0.9$ and $0.99$ for $\beta_1$ and $\beta_2$.
        
\section{Experimental Design}
\label{sec:ExperimentalDesign}
    This section describes the data set used for our experiments, which performance metrics were used, and how the evaluation was conducted.

    \subsection{Data}
    \label{sec:ED:Data}
        We evaluate the effect of our pre-training technique on the publicly available NASA \ac{cmapss} dataset.
        It contains four subsets (FD001 - FD004) of different operating conditions and possible fault modes.
        Engines in FD001 and FD002 experience only \ac{hpc} degradation, while the ones in FD003 and FD004 can additionally experience fan degradation.
        Further, in FD001 and FD003 are engines run under only one operating condition, while engines from FD002 and FD004 vary between six.
        Each subset is split into training and test data by the dataset authors.
        The training data contains multivariate time series data of aero engines up until the time step they failed.
        Each time series can be considered as a different engine of the same type.
        The test data's time series stop at a random time step before failure, for which the true \ac{rul} label is provided.
        Table \ref{tab:cmapss} summarizes the details of the dataset.
        We construct one validation set from each training set by taking $20\%$ of the time series.
        
        \begin{table}
        \centering
        \begin{tabular}{lcccc}
            \textbf{Dataset}                & \textbf{FD001} & \textbf{FD002} & \textbf{FD003} & \textbf{FD004}  \\
            \hline \hline
            \# Training Engines     & 100   & 260   & 100   & 249   \\
            \# Test Engines         & 100   & 259   & 100   & 248   \\
            \# Operation Conditions & 1     & 6     & 1     & 6     \\
            \# Failure Modes        & 1     & 1     & 2     & 2     \\
            \hline
        \end{tabular}
        \caption{\ac{cmapss} Dataset}
        \label{tab:cmapss}
        \end{table}
        
        For pre-processing, we follow Li et al. \cite{Li2018a} and select 14 of the 21 sensor channels with the indices 2, 3, 4, 7, 8, 9, 11, 12, 13, 14, 15, 17, 20, and 21 as the input of the network.
        The features are scaled by channel, using a min-max scaling calculated on each subsets' training set.
        We then use a sliding window with a step size of one to obtain frames of unified length in the time dimension.
        We use a window size of 30, 20, 30, and 15 for the subsets 1-4.
        These sizes were determined by the length shortest time series in each subsets test set.
        The \ac{rul} labels are calculated by a piece-wise linear degradation model with a maximum value $\textsc{RUL}_{\textsc{max}}$ of $125$.
        
    \subsection{Data Scenarios}
    \label{sec:ED:DataScenarios}
        As described in the introduction, unlabeled \ac{rul} data cannot contain the point of failure as we could simply compute the labels otherwise.
        It follows that we have to truncate the time series before failure when studying \ac{ssl}, as the model would have access to more failure data than normally available otherwise.
        How early we truncate the time series represents the \textit{grade of degradation} of the engine that the time series was collected from.
        Previous work on \ac{ssl} for \ac{rul} estimation varied only the amount of labeled data available, i.e. how many engines had already failed.
        Our experimental design includes varying the \textit{grade of degradation} for the unlabeled data, too.
    
        Adapting the work of Krokotsch et al. \cite{Krokotsch2020}, we impose data scenarios on each subset.
        In our case, a data scenario is characterized by the \textit{number of failed engines} and \textit{grade of degradation} of the unlabeled ones.
        Both factors are interpreted as percentages.
        A data scenario of \textit{number of engines} at $n\%$ would mean that only $n\%$ of the machines in the subset are used as labeled data for training.
        The rest is assumed to have not yet failed and is used as unlabeled data.
        This limits the amount of machine-to-machine variance that is covered by the labeled data.
        A data scenario of \textit{grade of degradation} at $n\%$ would mean that only the first $n\%$ of time steps of each unlabeled time series is available during training.
        This effectively limits the amount of available data near failure.
        
        We use five different \textit{grades of degradation} of $40\%$, $60\%$, $70\%$, $80\%$ and $90\%$ for our evaluation.
        A \textit{grades of degradation} of $100\%$ would not make sense, as the unlabeled machines would have failed already and could be used as labeled data.
        The lower limit of percentages is chosen due to the piece-wise linear degradation model.
        Using fewer than $40\%$ of the time steps would mean that the unlabeled data contains next to no data with a \ac{rul} of less than $\textsc{RUL}_{\textsc{max}}$.
        This means that the unlabeled data would come from completely healthy machines and may add no benefit to training.
        
        The number of failed engines is set at $2\%$, $10\%$, $20\%$, $40\%$, $100\%$.
        Using $100\%$ of failed machines means that no unlabeled data is available and pre-training is conducted on the labeled data only.
        We decided on these percentages because preliminary experiments did not show any degradation in performance for more than $40\%$ compared to using $100\%$ of the engines.
        The lower limit of $2\%$ is chosen because it results in at least one failed engine for each subset.

    \subsection{Performance Metrics}
    \label{sec:ED:Metrics}
        We employ two common performance metrics from the field of \ac{pdm}.
        The first is the \ac{rmse}, as described in equation (\ref{eqn:rmse}) over the test set.
        The second is the \emph{\ac{rul}-Score}, first proposed in the PHM 2008 Data Challenge \cite{Heimes2008}.
        It is calculated as follows:
        \begin{equation}
            s_i = 
            \begin{cases}
                e^{-\frac{\Delta\textsc{RUL}_i}{13}} - 1 & \text{for } \Delta\textsc{RUL}_i < 0 \\
                e^{\frac{\Delta\textsc{RUL}_i}{10}} - 1 & \text{for } \Delta\textsc{RUL}_i \geq 0
            \end{cases}
        \end{equation}
        where $\Delta\textsc{RUL}_i$ is the difference between predicted and true \ac{rul} for sample $x_i$.
        This metric penalizes overestimating \ac{rul} more than underestimating it as the former has a bigger impact on maintenance actions.
        We report the sum over all samples in the test set following previous work in the field.
        
        We will discuss the results using both metrics.
        Although, we will focus on \ac{rmse} as we view it as the more intuitive measure of performance.
    
    \subsection{Supervised Baseline}
    \label{sec:ED:Baseline}
        The baseline for our experiments is training the \ac{rul} estimation network in a supervised fashion as described in section \ref{sec:M:RULEstimation}.
        The training will only incorporate the available labeled data.
        We used the Adam optimizer \cite{Kingma2014a} with the default values for $0.9$ and $0.99$ for $\beta_1$ and $\beta_2$.
    
    \subsection{Competing Approaches}
    \label{sec:ED:Competition}
        Aside from the baseline, we compare our pre-training approach to two other methods found in the literature.
        The first approach is unsupervised pre-training via a deep \ac{ae}.
        We construct the \ac{ae} so that our feature extractor network is used as the encoder.
        The decoder is an exact mirror image of the encoder with transposed convolutions replacing the convolutional layers.
        The \ac{ae} is then trained to reconstruct its input via mini-batch gradient descent with a \ac{mse} loss.
        As for the baseline, we used the Adam optimizer.
        
        The second approach is unsupervised pre-training via a \ac{rbm}.
        The first layer of the feature extractor is interpreted as a \ac{rbm} with Gaussian visible units and \ac{relu} hidden units and trained until convergence.
        The other layers of the feature extractor remain randomly initialized.
        Again, Adam was used as the optimizer.
        
        Even though \cite{ListouEllefsen2019,Yoon2017} reported results for \acp{rbm} and variational \acp{ae}, we choose to reproduce the competing approaches ourselves.
        This has several reasons.
        First, the mentioned papers evaluated their approaches only on FD001 or FD004 which paints a limited picture of the approaches' performance.
        Second, the mentioned papers used a slightly different experimental setup, i.e. a different $\textsc{RUL}_{max}$, making comparison difficult.
        Furthermore, the aforementioned $\textsc{RUL}_{max}$ was optimized as a hyperparameter in one of the papers.
        As this value controls the possible range of the performance metrics, optimizing it results in improperly optimistic results.
    
    \subsection{Evaluation Procedure}
    \label{sec:ED:Evaluation}
        The baseline, our approach and the competing \ac{ssl} methods are evaluated on each of the subsets of the \ac{cmapss} dataset subject to each of our selected data scenarios.
        This results in 100 different experimental setups (4 subsets, 5 number of failed engines, 5 grades of degradation) for each \ac{ssl} method.
        The baseline includes only 20 setups, as it does not use unlabeled data which makes varying the grade of degradation superfluous.
        
        Each experimental setup is replicated ten times.
        For each replication, a new split of labeled and unlabeled data is randomly samples according to the data scenario.
        The splits are held constant across methods, i.e. the baseline receives the same labeled data as the \ac{ssl} methods given the same data scenario.
        This makes comparison easier as much more replications than ten would be needed for the low labeled data scenarios to receive statistically stable performance estimates.
        
        The pre-training stage for the \ac{ssl} methods receives the labeled and unlabeled portions of the data and trains for at least 100 epochs.
        The \ac{rbm} is trained for five epochs as it contains only one layer.
        Early stopping is used to select the model with the lowest validation loss.
        For our method we monitor the \ac{mse} loss for the relative \ac{rul} target and for the autoencoder the \ac{mse} reconstruction loss.
        
        The supervised training stage receives only the labeled portion of the data and is trained for at least 200 epochs.
        The weights of the selected pre-trained model are used to initialize the network in this stage for the \ac{ssl} methods.
        The baseline is initialized randomly.
        It should be noted that we divide the output of the feature extractor by its norm if the pre-training was self-supervised.
        Again, early stopping on the validation \ac{rmse} is used to select the best model for which the performance metrics over the test set are calculated.
        We report the performance for each subset/data scenario combination separately and averaged over the ten replications.

    \subsection{Hyperparameter Selection}
    \label{sec:ED:Hyperparameters}
        We began hyperparameter selection using the fixed network architecture shown in figure \ref{fig:BasicArchitecture} and conducted a random search in two steps.
        First, the hyperparameters of the supervised stage (i.e. learning rate, dropout, and batch size) were optimized for each subset of \ac{cmapss}.
        A network was trained in a supervised fashion with all available labeled data as described in section \ref{sec:ED:Evaluation}.
        After 100 trials, the hyperparameters of the network with the best validation \ac{rmse} were selected.
        The hyperparameters in question can be seen in table \ref{tab:hp_supervised}.
        
        \begin{table}[]
        \begin{tabular}{llllll}
        \multirow{2}{*}{\textbf{Hyperparameter}} & \multirow{2}{*}{\textbf{Search Space}}         & \multicolumn{4}{l}{\textbf{Configuration for}}                    \\
                                                 &                                                & \textbf{FD001} & \textbf{FD002} & \textbf{FD003} & \textbf{FD004} \\ 
        \hline \hline
        \textbf{Learning Rate}                   & $\operatorname{qlogu}(1^{-4}, 1^{-1}, 5^{-5})$ & $0.0056$       & $0.0903$       & $0.095$        & $0.06635$      \\
        \textbf{Dropout}                         & $\operatorname{qu}(0.0, 0.5, 0.1)$             & $0.4$          & $0.3$          & $0.2$          & $0.0$          \\
        \textbf{Batch Size}                      & [64, 128, 256, 512]                            & 128            & 512            & 64             & 64             \\
        \hline
        \end{tabular}
        \caption{\textbf{Hyperparameters of the Supervised Training Stage:} The search space $\operatorname{qlogu}(a,b,c)$ draws samples uniformly on a logarithmic scale from the interval $[a,b]$ quantized to $c$. The search space $\operatorname{qu}(a,b,c)$ draws samples uniformly from the interval $[a,b]$ quantized to $c$.}
        \label{tab:hp_supervised}
        \end{table}
        
        In a second step, the hyperparameters of the pre-training stages (i.e. learning rate, dropout, batch size, and minimum pair distance) were optimized similarly.
        The networks are trained without labels at $80\%$ grade of degradation as described in section \ref{sec:ED:Evaluation}.
        For our method we selected hyperparameters according to the validation \ac{mse} loss for the relative \ac{rul} target and for the autoencoder according to the validation \ac{mse} reconstruction loss.
        As stability was an issue, each hyperparameter configuration was trained five times and the mean of these replications is used for selection.
        For the \ac{rbm} we adopted the hyperparameters from \cite{ListouEllefsen2019}.
        We set the learning rate for this method to $10^{-4}$ by hand as it was not given in the paper.
        
        It should be noted that all optimizations used no labels at all and can therefore be conducted independently from the amount of labeled data.
        The selected hyperparameters can be seen in table \ref{tab:hp_self_supervised} and \ref{tab:hp_unsupervised}.
        
        \begin{table}[]
        \begin{tabular}{llllll}
        \multirow{2}{*}{\textbf{Hyperparameter}} & \multirow{2}{*}{\textbf{Search Space}}         & \multicolumn{4}{l}{\textbf{Configuration for}}                    \\
                                                 &                                                & \textbf{FD001} & \textbf{FD002} & \textbf{FD003} & \textbf{FD004} \\
        \hline \hline
        \textbf{Learning Rate}                   & $\operatorname{qlogu}(1^{-4}, 1^{-1}, 5^{-5})$ & $0.00015$      & $0.01155$      & $0.00615$      & $0.07455$      \\
        \textbf{Dropout}                         & $\operatorname{qu}(0.0, 0.5, 0.1)$             & $0.2$          & $0.4$          & $0.1$          & $0.1$          \\
        \textbf{Batch Size}                      & [64, 128, 256, 512]                            & 64             & 64             & 64             & 64             \\
        \textbf{Minimum Distance}                & [1, 10, 15, 30]                                & 10             & 15             & 15             & 10             \\
        \hline
        \end{tabular}
        \caption{\textbf{Hyperparameters of the Self-Supervised Pre-Training:} See table \ref{tab:hp_supervised} for an explanation of the search spaces.}
        \label{tab:hp_self_supervised}
        \end{table}
        
        \begin{table}[]
        \begin{tabular}{llllll}
        \multirow{2}{*}{\textbf{Hyperparameter}} & \multirow{2}{*}{\textbf{Search Space}}         & \multicolumn{4}{l}{\textbf{Configuration for}}                    \\
                                                 &                                                & \textbf{FD001} & \textbf{FD002} & \textbf{FD003} & \textbf{FD004} \\
        \hline \hline
        \textbf{Learning Rate}                   & $\operatorname{qlogu}(1^{-4}, 1^{-1}, 5^{-5})$ & $0.0001$      & $0.0248$      & $0.015$      & $0.0006$      \\
        \textbf{Dropout}                         & $\operatorname{qu}(0.0, 0.5, 0.1)$             & $0.1$          & $0.4$          & $0.0$          & $0.0$          \\
        \textbf{Batch Size}                      & [64, 128, 256, 512]                            & 64             & 256             & 64             & 64             \\
        \textbf{Minimum Distance}                & [1, 10, 15, 30]                                & 1             & 15             & 1             & 10             \\
        \hline
        \end{tabular}
        \caption{\textbf{Hyperparameters of the Autoencoder Pre-Training:} See table \ref{tab:hp_supervised} for an explanation of the search spaces. The minimum distance was optimized for this method, too, due to implementation reasons. It should not influence the results significantly, though.}
        \label{tab:hp_unsupervised}
        \end{table}
    
\section{Results}
\label{sec:Results}
    First, we will describe the results of our experiments for each \ac{cmapss} subset.
    Afterward, we will interpret the findings and set them into context.

    \subsection{Comparison of Approaches}
    \label{sec:R:Comparison}
\begin{figure}
        \centering
        \includegraphics[width=\textwidth]{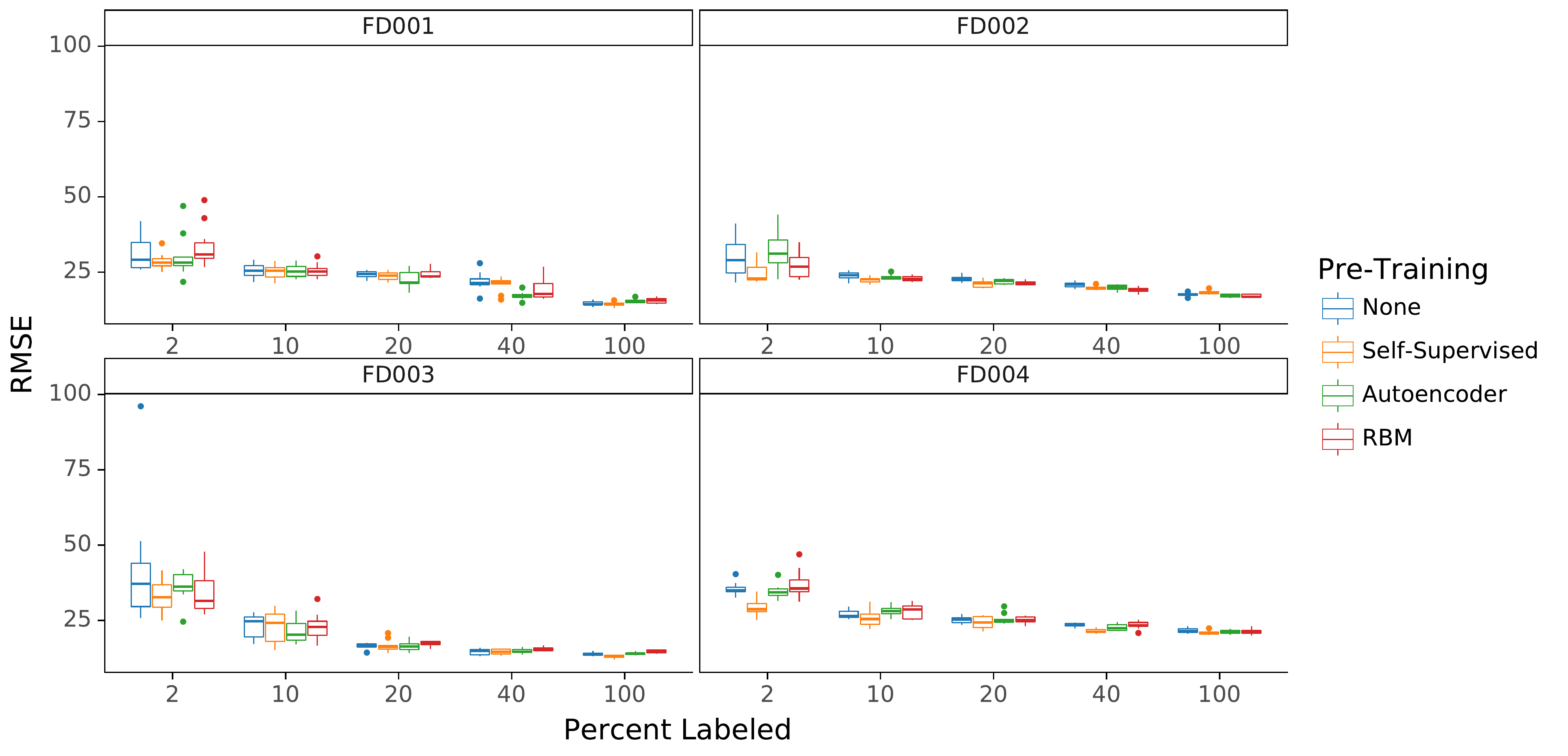}
        \caption{\textbf{\ac{rmse} at $\mathbf{80\%}$ degradation:} The plots show the results on the four subsets of the \ac{cmapss} dataset. We plot the \ac{rmse} test performance against the percentage of labeled data. The remaining data were used as unlabeled data with a grade of degradation of $80\%$. The box represents the \ac{iqr} and the middle line the median performance. The ends of the whiskers depict the minimum and maximum performance that was not deemed an outlier. Outliers are defined as more than 1.5 times \ac{iqr} away from the lower or upper end of the box. We can see that the performance degrades only slowly with less labeled data. Significant drops in performance can be seen for $10\%$ labeled data for FD003, 40\% for FD001, and $2\%$ for the other subsets.}
        \label{fig:results_80_degradation}
    \end{figure}

\begin{figure}
        \centering
        \includegraphics[width=\textwidth]{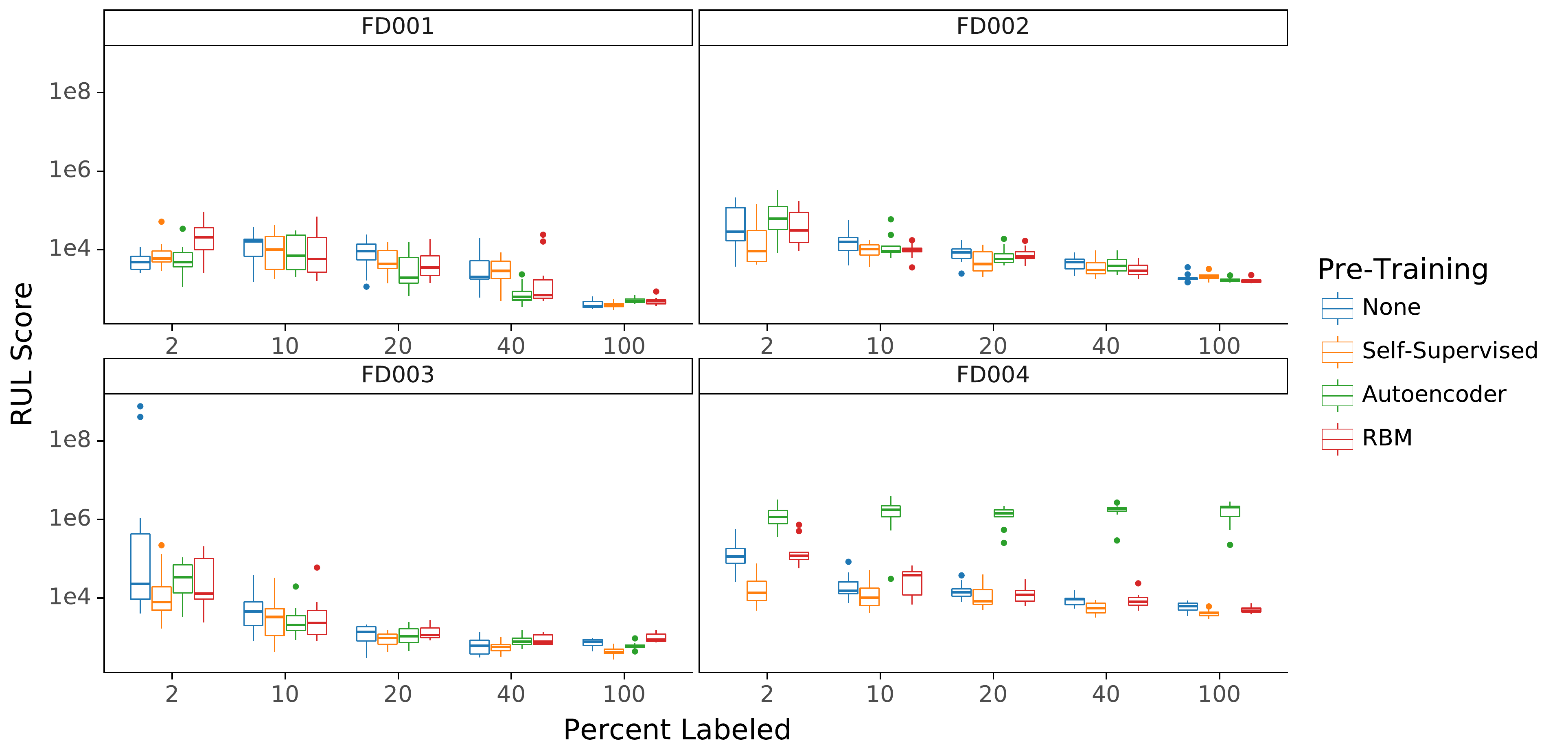}
        \caption{\textbf{\ac{rul}-Score at $\mathbf{80\%}$ degradation:} The plots show the results on the four subsets of the \ac{cmapss} dataset. We plot the \ac{rul} Score test performance on a logarithmic scale against the percentage of labeled data. The remaining data was used as unlabeled data with a grade of degradation of $80\%$. See figure \ref{fig:results_80_degradation} for an explanation of the box plots. The results are similar to the \ac{rmse} in figure \ref{fig:results_80_degradation} but show significantly worse scores for the \ac{ae} pre-training on FD004.}
        \label{fig:results_score_80_degradation}
    \end{figure}

\begin{figure}
        \centering
        \includegraphics[width=\textwidth]{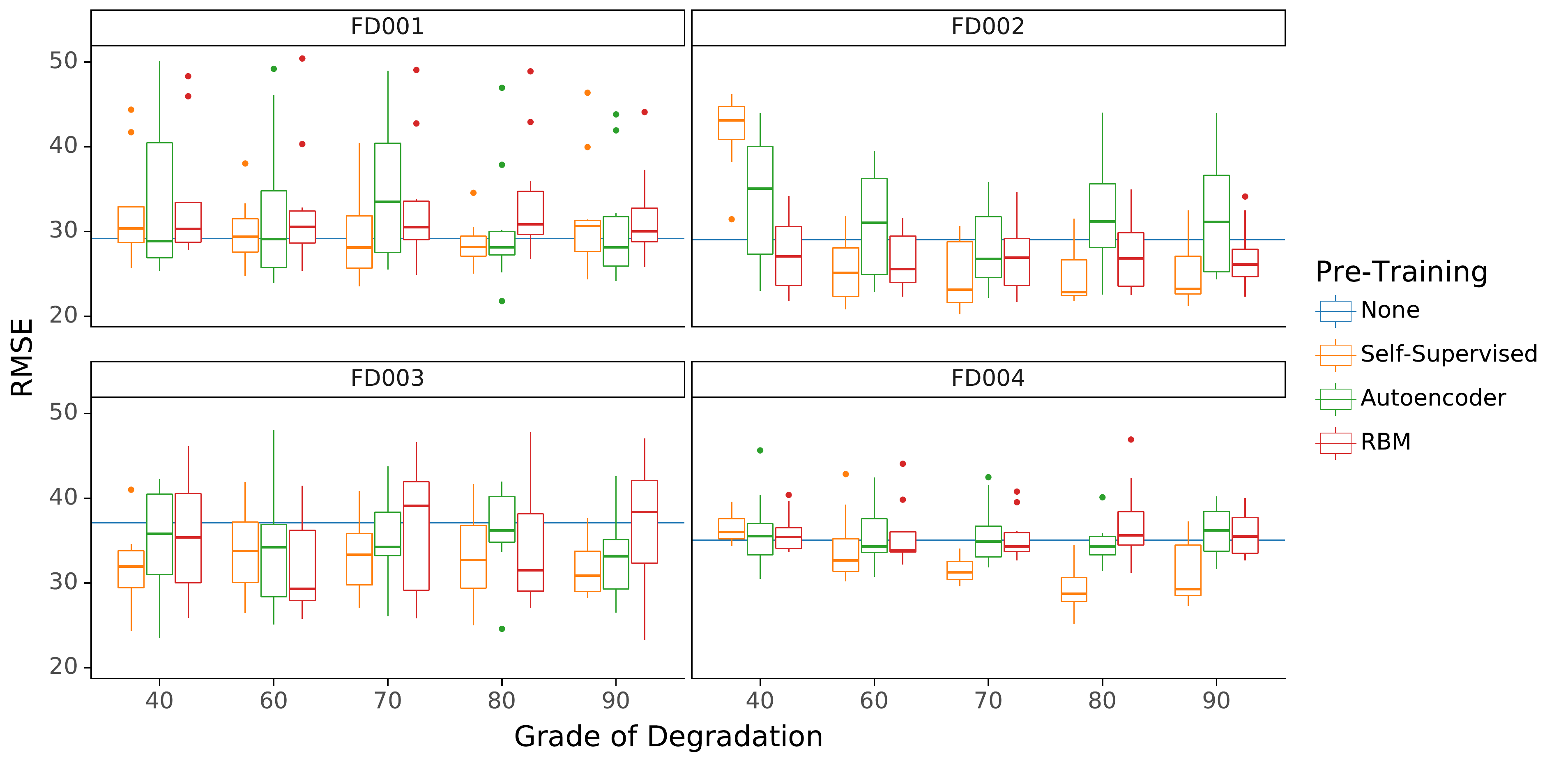}
        \caption{\textbf{\ac{rmse} at $\mathbf{2\%}$ of labeled data:} The plots show the results on the four subsets of the \ac{cmapss} dataset. We plot the \ac{rmse} test performance against the grade of degradation when only $2\%$ of the labeled data was used. The median performance of the baseline is depicted in blue. See figure \ref{fig:results_80_degradation} for an explanation of the box plots. We can observe that the performance of the \ac{ae} and the \ac{rbm} is relatively constant on all grades of degradation. The self-supervised method's performance degrades for lower degradation, which can be seen best in FD004. Nevertheless, the self-supervised method shows much better performance on higher grades of degradation in FD002 and FD004 than the competing methods. On the remaining subsets, the performance of the self-supervised method is comparable to the other approaches.}
        \label{fig:results_2_labeled}
    \end{figure}

\begin{figure}
        \centering
        \includegraphics[width=\textwidth]{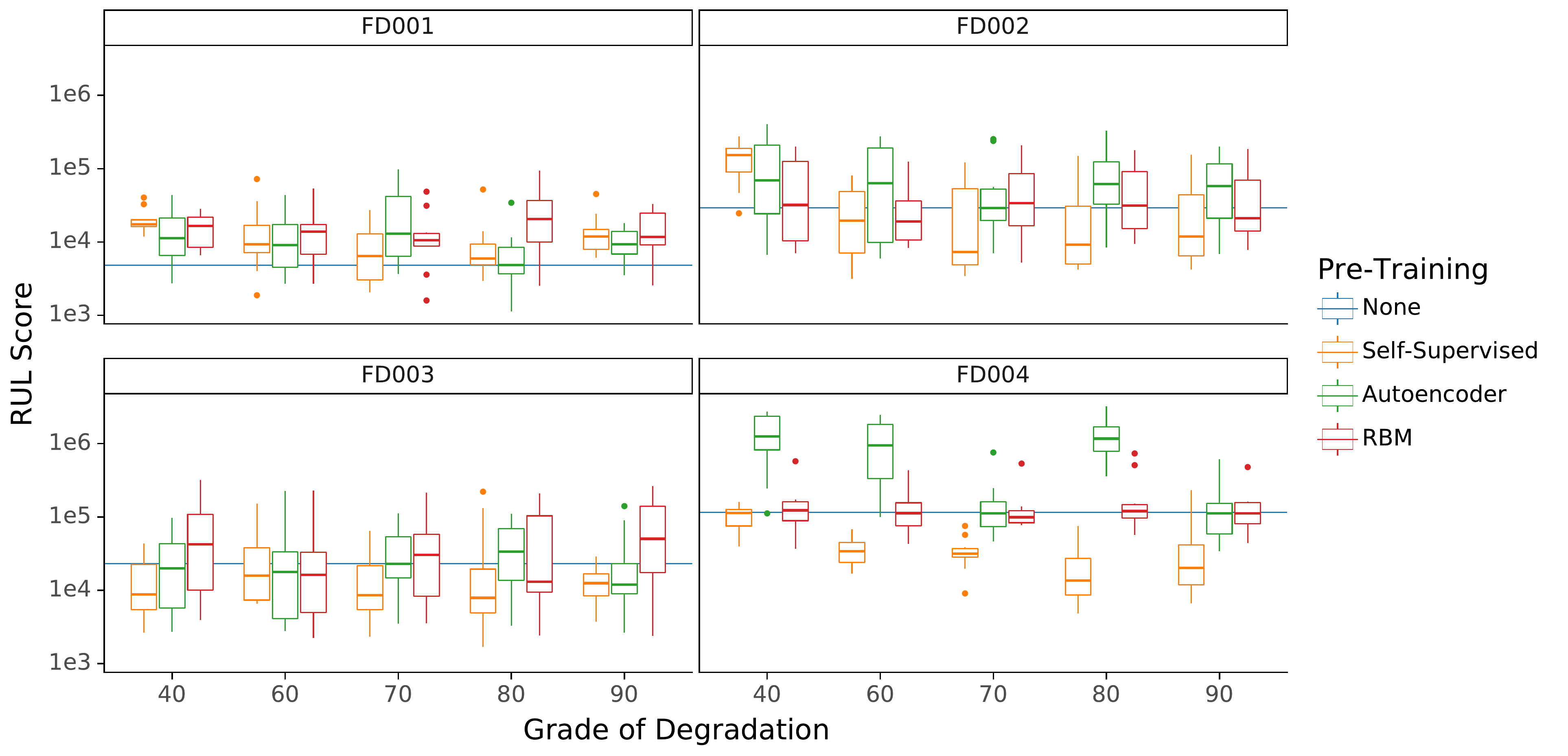}
        \caption{\textbf{\ac{rul}-Score at $\mathbf{2\%}$ of labeled data:} The plots show the results on the four subsets of the \ac{cmapss} dataset. We plot the \ac{rul} Score test performance on a logarithmic scale against the grade of degradation when only $2\%$ of the labeled data was used. The median performance of the baseline is depicted in blue. See figure \ref{fig:results_80_degradation} for an explanation of the box plots. The results are similar to the \ac{rmse} in figure \ref{fig:results_80_degradation} but show significantly worse scores for the \ac{ae} pre-training on FD004. We can also observe that the median score of the self-supervised approach is the lowest for all grades of degradation on FD002. FD003 and FD004.}
        \label{fig:results_score_2_labeled}
    \end{figure}

        Our experiments produced too many data points to present them all in detail.
        We will therefore show only slices of our results as plots.
        The results plotted against the percentage of labeled data are shown in figures \ref{fig:results_80_degradation} and \ref{fig:results_score_80_degradation} at 80\% degradation.
        The results plotted against grade of degradation are shown in figures \ref{fig:results_2_labeled} and \ref{fig:results_score_2_labeled} at 2\% of the labeled data.
        The complete results are shown in tables \ref{tab:res_12}, \ref{tab:res_34}, \ref{tab:score_12} and \ref{tab:score_34}.
        
        Overall we can conclude that a significant drop in baseline performance was mostly observable for very low amounts of labeled data.
        In figure \ref{fig:results_80_degradation} we can see that the performance of the baseline (blue) has a relatively small standard deviation for 100\%, 40\%, and 20\% of labeled data.
        For the subsets FD002 and FD004, the mean and standard deviation increase only at 2\% labeled data.
        For FD001, the performance already drops at 40\% labeled data and for FD003 at 10\%.
        One has to keep in mind, though, that FD002 and FD004 are more than twice as large as FD001 and FD003, which means that they have a higher amount of labeled data available at the same percentage.
        
        Performance on \textbf{FD001} shows next to no improvement through \ac{ssl}.
        In figure \ref{fig:results_80_degradation} we can see that the median \ac{rmse} performances is well inside of each others \acp{iqr} with the exception of 40\% labeled data where \ac{ae} pre-training achieves much better performance.
        In one case, i.e. for 2\% labeled data, the performance of \ac{rbm} pre-training was even worse than the baseline.
        The \ac{rul}-Score metric in figure \ref{fig:results_score_80_degradation} paints a similar picture, even though pre-trained models seem slightly better for 20\% and 40\% labeled data.
        For 2\% labeled data, the \ac{ssl} approaches had a worse mean \ac{rul}-Score performance than the baseline.
        These findings are stable for different grades of degradation, too.
        Figures \ref{fig:results_2_labeled} and \ref{fig:results_score_2_labeled} show no discernible trend with respect to the grade of degradation for any method.
        Overall, the \ac{ae} seems to perform slightly better than the other approaches.
        
        Performance on \textbf{FD002} clearly benefits from \ac{rbm} and self-supervised pre-training.
        The self-supervised pre-training, in particular, beats the baseline and the other approaches most of the time, especially in low-labeled data scenarios.
        On the other hand, there is an extreme drop in performance for our method at 40\% degradation.
        This is true for all percentages of labeled data but most apparent for 2\%.
        When looking at figure \ref{fig:results_score_2_labeled}, only self-supervised pre-training reliably beats the median performance of the baseline in terms of \ac{rul}-Score for grades of degradation above 40\%.
        
        Performance gains on \textbf{FD003} are better than on FD001 but still minor.
        For 2\% labeled data, the self-supervised pre-training beats the baseline and the other methods for next to all grades of degradation in terms of mean \ac{rmse} and \ac{rul}-score.
        Nevertheless, the \acp{iqr} of our method and the \ac{rbm} are often highly overlapping.
        As on FD001, we can see no trend concerning the grade of degradation.
        
        On \textbf{FD004}, we can see the benefits of our approach most clearly.
        While \ac{ae} and \ac{rbm} pre-training bring next to no performance gains in terms of median \ac{rmse}, self-supervised pre-training outperforms the baseline reliably, as seen in figure \ref{fig:results_80_degradation} and \ref{fig:results_2_labeled}.
        Nevertheless, we can observe a downward trend in performance with respect to the grade of degradation for our method.
        The competing approaches do not suffer from this.
        Figures \ref{fig:results_score_80_degradation} and \ref{fig:results_score_2_labeled} reveal devastating performance losses for \ac{ae} pre-training in terms of \ac{rul}-Score.
        
        We can conclude that \ac{ssl} can be beneficial for \ac{rul} estimation even under realistic conditions with varying grades of degradation.
        However, no approach was able to reliably beat the baseline on all subsets under all data scenarios.
        A representative validation set and careful hyperparameter tuning are needed to assure improved performance and detect negative outcomes.
        Self-supervised pre-training seems to be a step in the right direction, as it often outperforms the competing approaches.
        Nevertheless, its performance trends downward on FD002 and FD004 with a falling grade of degradation.
    
    \begin{table}
    \begin{adjustbox}{width=\textwidth,center}
    \centering
    \begin{tabular}{cl|ccccc}
    \hline
        &   & \multicolumn{5}{c}{FD001}   \\
                     &    & 2\%     & 10\%    & 20\%    & 40\%    & 100\%   \\
    \hline \hline
                           & None       & $31.30 \pm 6.16$          & $25.45 \pm 2.31$          & $24.21 \pm 1.31$          & $21.94 \pm 3.08$          & $14.63 \pm 0.81$          \\
    \hline
     \multirow{3}{*}{40\%} & AE         & $33.37 \pm 9.53$          & $24.71 \pm 1.28$          & $\mathbf{23.56 \pm 3.12}$ & $\mathbf{17.27 \pm 1.47}$ & $15.03 \pm 0.61$          \\
                           & RBM        & $33.49 \pm 7.49$          & $25.90 \pm 1.93$          & $25.09 \pm 1.61$          & $21.02 \pm 3.17$          & $15.67 \pm 0.57$          \\
                           & Ours       & $32.35 \pm 6.08$          & $\mathbf{24.42 \pm 1.58}$ & $24.91 \pm 1.87$          & $21.38 \pm 2.32$          & $\mathbf{14.34 \pm 0.81}$ \\
    \hline
     \multirow{3}{*}{60\%} & AE         & $32.17 \pm 8.94$          & $\mathbf{24.62 \pm 1.75}$ & $\mathbf{22.65 \pm 2.46}$ & $\mathbf{17.95 \pm 1.92}$ & $15.09 \pm 0.52$          \\
                           & RBM        & $32.64 \pm 7.45$          & $27.06 \pm 1.79$          & $24.78 \pm 2.73$          & $19.50 \pm 2.93$          & $15.68 \pm 1.01$          \\
                           & Ours       & $\mathbf{29.88 \pm 3.83}$ & $25.81 \pm 2.36$          & $23.93 \pm 1.72$          & $22.41 \pm 1.82$          & $\mathbf{14.48 \pm 0.87}$ \\
    \hline
     \multirow{3}{*}{70\%} & AE         & $34.93 \pm 8.64$          & $25.10 \pm 2.23$          & $\mathbf{21.70 \pm 2.09}$ & $\mathbf{17.71 \pm 1.79}$ & $15.02 \pm 0.58$          \\
                           & RBM        & $33.00 \pm 7.40$          & $26.00 \pm 1.77$          & $23.95 \pm 1.55$          & $19.21 \pm 1.92$          & $15.45 \pm 0.71$          \\
                           & Ours       & $\mathbf{29.52 \pm 5.26}$ & $\mathbf{24.90 \pm 1.47}$ & $22.96 \pm 2.38$          & $18.51 \pm 2.76$          & $15.35 \pm 0.78$          \\
    \hline
     \multirow{3}{*}{80\%} & AE         & $30.24 \pm 7.16$          & $25.32 \pm 2.23$          & $\mathbf{22.41 \pm 3.07}$ & $\mathbf{17.13 \pm 1.31}$ & $15.40 \pm 0.62$          \\
                           & RBM        & $33.58 \pm 7.05$          & $25.53 \pm 2.36$          & $24.43 \pm 1.66$          & $19.56 \pm 4.02$          & $15.55 \pm 0.88$          \\
                           & Ours       & $\mathbf{28.50 \pm 2.68}$ & $\mathbf{25.13 \pm 2.28}$ & $23.64 \pm 1.52$          & $20.98 \pm 2.52$          & $\mathbf{14.33 \pm 0.74}$ \\
    \hline
     \multirow{3}{*}{90\%} & AE         & $\mathbf{30.57 \pm 7.00}$ & $26.04 \pm 1.92$          & $\mathbf{21.84 \pm 1.82}$ & $\mathbf{17.26 \pm 1.34}$ & $14.89 \pm 0.65$          \\
                           & RBM        & $31.86 \pm 5.30$          & $\mathbf{25.05 \pm 2.17}$ & $25.01 \pm 2.53$          & $19.44 \pm 2.36$          & $15.55 \pm 0.47$          \\
                           & Ours       & $31.69 \pm 6.62$          & $25.20 \pm 1.92$          & $23.52 \pm 2.24$          & $18.74 \pm 3.53$          & $15.39 \pm 0.78$          \\
    \hline
    \hline
        &   & \multicolumn{5}{c}{FD002}   \\
                     &    & 2\%     & 10\%    & 20\%    & 40\%    & 100\%   \\
    \hline \hline
                           & None       & $29.80 \pm 6.38$          & $23.87 \pm 1.30$          & $22.83 \pm 1.03$          & $20.73 \pm 0.94$          & $17.55 \pm 0.62$          \\
    \hline
     \multirow{3}{*}{40\%} & AE         & $33.86 \pm 7.88$          & $23.12 \pm 1.11$          & $21.91 \pm 0.96$          & $20.48 \pm 0.90$          & $\mathbf{17.19 \pm 0.30}$ \\
                           & RBM        & $\mathbf{27.43 \pm 4.30}$ & $\mathbf{22.79 \pm 0.70}$ & $\mathbf{20.98 \pm 0.65}$ & $\mathbf{19.41 \pm 0.92}$ & $17.19 \pm 0.51$          \\
                           & Ours       & $42.04 \pm 4.47$          & $24.04 \pm 0.84$          & $22.09 \pm 1.31$          & $20.44 \pm 0.80$          & $18.34 \pm 0.84$          \\
    \hline
     \multirow{3}{*}{60\%} & AE         & $30.95 \pm 6.38$          & $23.25 \pm 0.92$          & $22.26 \pm 1.05$          & $20.69 \pm 0.91$          & $17.78 \pm 1.22$          \\
                           & RBM        & $26.50 \pm 3.22$          & $22.87 \pm 0.90$          & $\mathbf{21.36 \pm 1.01}$ & $\mathbf{19.42 \pm 0.68}$ & $\mathbf{17.54 \pm 0.74}$ \\
                           & Ours       & $\mathbf{25.46 \pm 3.79}$ & $\mathbf{22.86 \pm 1.17}$ & $22.14 \pm 1.35$          & $20.45 \pm 0.84$          & $18.45 \pm 0.78$          \\
    \hline
     \multirow{3}{*}{70\%} & AE         & $28.03 \pm 4.85$          & $23.84 \pm 1.06$          & $22.43 \pm 1.83$          & $20.49 \pm 0.82$          & $17.56 \pm 1.22$          \\
                           & RBM        & $26.83 \pm 3.91$          & $22.85 \pm 1.20$          & $21.41 \pm 1.19$          & $19.33 \pm 0.58$          & $17.17 \pm 0.46$          \\
                           & Ours       & $\mathbf{24.82 \pm 4.00}$ & $\mathbf{21.79 \pm 1.42}$ & $\mathbf{20.25 \pm 0.77}$ & $\mathbf{18.66 \pm 0.69}$ & $\mathbf{16.83 \pm 0.44}$ \\
    \hline
     \multirow{3}{*}{80\%} & AE         & $32.36 \pm 7.24$          & $23.29 \pm 0.90$          & $21.87 \pm 0.90$          & $19.92 \pm 0.90$          & $17.21 \pm 0.58$          \\
                           & RBM        & $27.13 \pm 4.05$          & $22.84 \pm 0.86$          & $21.35 \pm 0.66$          & $\mathbf{19.13 \pm 0.87}$ & $\mathbf{17.10 \pm 0.57}$ \\
                           & Ours       & $\mathbf{24.82 \pm 3.43}$ & $\mathbf{22.43 \pm 1.11}$ & $\mathbf{21.21 \pm 1.26}$ & $19.79 \pm 0.65$          & $18.23 \pm 0.63$          \\
    \hline
     \multirow{3}{*}{90\%} & AE         & $31.55 \pm 6.75$          & $23.70 \pm 1.07$          & $22.30 \pm 1.00$          & $20.71 \pm 1.36$          & $17.55 \pm 0.53$          \\
                           & RBM        & $26.98 \pm 3.73$          & $22.55 \pm 1.12$          & $20.99 \pm 1.07$          & $19.46 \pm 0.70$          & $17.28 \pm 0.39$          \\
                           & Ours       & $\mathbf{24.99 \pm 3.80}$ & $\mathbf{21.48 \pm 1.39}$ & $\mathbf{20.26 \pm 1.23}$ & $\mathbf{18.82 \pm 0.72}$ & $\mathbf{16.80 \pm 0.44}$ \\
    \hline
    \end{tabular}
    \end{adjustbox}
    \caption{\textbf{\ac{rmse} results for FD001 and FD002}: We report the mean and standard deviation. The rows represent the \emph{grade of degradation} and the columns the \emph{percent of labeled data} of the data scenario. The second column contains the pre-training method, where \emph{None} is the baseline without any pre-training, and \emph{Ours} is the self-supervised pre-training. The bold results mark the best mean performance for each data scenario. If no result is bold, no approach was able to beat the baseline.}
    \label{tab:res_12}
    \end{table}
    
    \begin{table}
    \begin{adjustbox}{width=\textwidth,center}
    \centering
    \begin{tabular}{cl|ccccc}
    \hline
        &   & \multicolumn{5}{c}{FD003}   \\
                     &    & 2\%     & 10\%    & 20\%    & 40\%    & 100\%   \\
    \hline \hline
                          & None       & $42.33 \pm 20.56$         & $22.99 \pm 4.02$          & $16.55 \pm 0.94$          & $14.58 \pm 1.06$          & $13.88 \pm 0.58$          \\
    \hline
     \multirow{3}{*}{40\%} & AE         & $35.08 \pm 6.38$          & $\mathbf{22.12 \pm 5.31}$ & $17.05 \pm 1.63$          & $14.75 \pm 1.12$          & $13.96 \pm 0.59$          \\
                           & RBM        & $35.83 \pm 7.00$          & $22.98 \pm 4.85$          & $16.82 \pm 1.13$          & $15.19 \pm 0.82$          & $14.50 \pm 0.50$          \\
                           & Ours       & $\mathbf{31.82 \pm 4.51}$ & $22.98 \pm 4.30$          & $\mathbf{16.00 \pm 0.86}$ & $\mathbf{14.39 \pm 0.85}$ & $\mathbf{13.47 \pm 0.18}$ \\
    \hline
     \multirow{3}{*}{60\%} & AE         & $34.00 \pm 6.89$          & $\mathbf{21.26 \pm 4.86}$ & $16.77 \pm 1.62$          & $15.09 \pm 1.14$          & $14.16 \pm 0.49$          \\
                           & RBM        & $\mathbf{31.63 \pm 6.02}$ & $22.63 \pm 3.93$          & $17.04 \pm 1.89$          & $15.33 \pm 0.84$          & $14.45 \pm 0.42$          \\
                           & Ours       & $33.79 \pm 4.74$          & $24.80 \pm 4.77$          & $17.10 \pm 1.19$          & $\mathbf{14.52 \pm 1.10}$ & $\mathbf{13.09 \pm 0.74}$ \\
    \hline
     \multirow{3}{*}{70\%} & AE         & $35.01 \pm 5.99$          & $\mathbf{22.00 \pm 4.80}$ & $\mathbf{16.36 \pm 1.15}$ & $15.39 \pm 0.93$          & $13.85 \pm 0.42$          \\
                           & RBM        & $36.30 \pm 7.52$          & $22.36 \pm 4.23$          & $16.68 \pm 1.35$          & $15.51 \pm 0.88$          & $14.61 \pm 0.45$          \\
                           & Ours       & $\mathbf{33.10 \pm 4.34}$ & $22.35 \pm 4.82$          & $16.73 \pm 1.45$          & $15.14 \pm 1.18$          & $\mathbf{13.64 \pm 0.62}$ \\
    \hline
     \multirow{3}{*}{80\%} & AE         & $36.35 \pm 5.06$          & $\mathbf{21.24 \pm 3.75}$ & $\mathbf{16.49 \pm 1.60}$ & $14.82 \pm 0.76$          & $14.03 \pm 0.50$          \\
                           & RBM        & $34.29 \pm 6.83$          & $22.87 \pm 4.63$          & $17.31 \pm 0.86$          & $15.52 \pm 0.66$          & $14.68 \pm 0.56$          \\
                           & Ours       & $\mathbf{33.47 \pm 5.57}$ & $22.98 \pm 5.18$          & $16.54 \pm 2.04$          & $14.60 \pm 0.91$          & $\mathbf{13.05 \pm 0.53}$ \\
    \hline
     \multirow{3}{*}{90\%} & AE         & $33.37 \pm 5.36$          & $\mathbf{20.14 \pm 4.09}$ & $17.10 \pm 2.03$          & $14.84 \pm 0.88$          & $14.17 \pm 0.35$          \\
                           & RBM        & $36.48 \pm 7.74$          & $23.47 \pm 4.27$          & $17.26 \pm 1.35$          & $15.06 \pm 0.92$          & $14.34 \pm 0.57$          \\
                           & Ours       & $\mathbf{31.79 \pm 3.53}$ & $21.19 \pm 4.13$          & $\mathbf{16.34 \pm 1.19}$ & $14.78 \pm 0.75$          & $\mathbf{13.83 \pm 0.49}$ \\
    \hline
    \hline
        &   & \multicolumn{5}{c}{FD004}   \\
                     &    & 2\%     & 10\%    & 20\%    & 40\%    & 100\%   \\
    \hline \hline
                           & None       & $35.44 \pm 2.22$          & $27.00 \pm 1.37$          & $25.18 \pm 1.13$          & $23.51 \pm 0.62$          & $21.66 \pm 0.90$          \\
    \hline
     \multirow{3}{*}{40\%} & AE         & $36.07 \pm 4.39$          & $28.53 \pm 2.26$          & $25.39 \pm 1.36$          & $\mathbf{23.20 \pm 1.06}$ & $21.62 \pm 0.71$          \\
                           & RBM        & $35.99 \pm 2.38$          & $28.46 \pm 2.20$          & $25.74 \pm 0.90$          & $23.55 \pm 0.62$          & $21.48 \pm 0.54$          \\
                           & Ours       & $36.41 \pm 1.71$          & $31.15 \pm 3.66$          & $25.44 \pm 1.67$          & $23.24 \pm 1.84$          & $\mathbf{21.11 \pm 0.52}$ \\
    \hline
     \multirow{3}{*}{60\%} & AE         & $35.60 \pm 3.68$          & $28.34 \pm 2.36$          & $\mathbf{24.88 \pm 0.97}$ & $23.42 \pm 0.97$          & $21.30 \pm 0.37$          \\
                           & RBM        & $35.59 \pm 3.69$          & $27.61 \pm 2.44$          & $25.48 \pm 1.22$          & $23.22 \pm 0.48$          & $21.18 \pm 0.62$          \\
                           & Ours       & $\mathbf{34.20 \pm 4.08}$ & $29.40 \pm 2.77$          & $25.31 \pm 2.28$          & $\mathbf{22.24 \pm 1.21}$ & $\mathbf{20.96 \pm 0.67}$ \\
    \hline
     \multirow{3}{*}{70\%} & AE         & $35.72 \pm 3.72$          & $28.28 \pm 1.82$          & $25.34 \pm 1.30$          & $23.19 \pm 0.71$          & $21.26 \pm 0.56$          \\
                           & RBM        & $35.41 \pm 2.69$          & $27.32 \pm 2.30$          & $25.40 \pm 1.39$          & $23.02 \pm 0.73$          & $21.08 \pm 0.52$          \\
                           & Ours       & $\mathbf{31.58 \pm 1.57}$ & $27.02 \pm 2.07$          & $\mathbf{23.88 \pm 2.14}$ & $\mathbf{22.48 \pm 1.51}$ & $\mathbf{20.95 \pm 0.67}$ \\
    \hline
     \multirow{3}{*}{80\%} & AE         & $34.68 \pm 2.34$          & $28.16 \pm 1.73$          & $25.47 \pm 1.79$          & $22.74 \pm 1.08$          & $21.23 \pm 0.62$          \\
                           & RBM        & $37.15 \pm 4.56$          & $27.95 \pm 2.36$          & $25.20 \pm 1.08$          & $23.43 \pm 1.25$          & $21.26 \pm 0.89$          \\
                           & Ours       & $\mathbf{29.23 \pm 2.74}$ & $\mathbf{25.84 \pm 2.84}$ & $\mathbf{24.26 \pm 2.11}$ & $\mathbf{21.46 \pm 0.68}$ & $\mathbf{20.95 \pm 0.67}$ \\
    \hline
     \multirow{3}{*}{90\%} & AE         & $35.95 \pm 2.96$          & $27.57 \pm 1.28$          & $25.66 \pm 1.14$          & $23.43 \pm 1.20$          & $21.73 \pm 0.59$          \\
                           & RBM        & $35.88 \pm 2.70$          & $27.84 \pm 1.82$          & $25.69 \pm 1.31$          & $23.73 \pm 0.78$          & $21.25 \pm 0.44$          \\
                           & Ours       & $\mathbf{31.07 \pm 3.81}$ & $\mathbf{24.20 \pm 2.14}$ & $\mathbf{24.17 \pm 2.88}$ & $\mathbf{21.82 \pm 1.22}$ & $\mathbf{20.90 \pm 0.62}$ \\
    \hline
    \end{tabular}
    \end{adjustbox}
    \caption{\textbf{\ac{rmse} results for FD003 and FD004}: Please consult table \ref{tab:res_12} for further information.}
    \label{tab:res_34}
    \end{table}
    
    \begin{table}
    \begin{adjustbox}{width=\textwidth,center}
    \centering
    \begin{tabular}{cl|ccccc}
    \hline
        &   & \multicolumn{5}{c}{FD001}   \\
                     &    & 2\%     & 10\%    & 20\%    & 40\%    & 100\%   \\
    \hline \hline
                           & None       & $5.64e3 \pm 3.25e3$ & $1.49e4 \pm 1.11e4$          & $1.04e4 \pm 7.48e3$          & $5.28e3 \pm 6.35e3$          & $4.24e2 \pm 1.15e2$          \\
    \hline
     \multirow{3}{*}{40\%} & AE         & $1.51e4 \pm 1.28e4$ & $\mathbf{4.57e3 \pm 2.37e3}$ & $\mathbf{7.39e3 \pm 7.22e3}$ & $\mathbf{1.03e3 \pm 7.22e2}$ & $4.79e2 \pm 1.03e2$          \\
                           & RBM        & $1.62e4 \pm 8.24e3$ & $1.73e4 \pm 1.83e4$          & $9.79e3 \pm 9.05e3$          & $3.69e3 \pm 4.48e3$          & $5.32e2 \pm 6.25e1$          \\
                           & Ours       & $2.04e4 \pm 9.05e3$ & $1.14e4 \pm 8.40e3$          & $1.57e4 \pm 1.32e4$          & $3.90e3 \pm 2.89e3$          & $\mathbf{3.97e2 \pm 8.31e1}$ \\
    \hline
     \multirow{3}{*}{60\%} & AE         & $1.38e4 \pm 1.28e4$ & $\mathbf{9.80e3 \pm 1.20e4}$ & $\mathbf{4.76e3 \pm 4.94e3}$ & $\mathbf{9.85e2 \pm 5.46e2}$ & $4.79e2 \pm 6.05e1$          \\
                           & RBM        & $1.76e4 \pm 1.56e4$ & $1.89e4 \pm 2.34e4$          & $1.55e4 \pm 1.58e4$          & $2.95e3 \pm 4.13e3$          & $5.14e2 \pm 9.48e1$          \\
                           & Ours       & $1.80e4 \pm 2.13e4$ & $1.08e4 \pm 5.36e3$          & $8.08e3 \pm 5.60e3$          & $6.04e3 \pm 5.29e3$          & $\mathbf{4.04e2 \pm 8.41e1}$ \\
    \hline
     \multirow{3}{*}{70\%} & AE         & $3.16e4 \pm 3.61e4$ & $1.42e4 \pm 1.67e4$          & $\mathbf{2.71e3 \pm 2.45e3}$ & $\mathbf{8.92e2 \pm 4.70e2}$ & $4.76e2 \pm 7.56e1$          \\
                           & RBM        & $1.49e4 \pm 1.42e4$ & $1.37e4 \pm 1.51e4$          & $5.71e3 \pm 5.43e3$          & $1.06e3 \pm 5.36e2$          & $4.91e2 \pm 1.05e2$          \\
                           & Ours       & $9.03e3 \pm 8.16e3$ & $\mathbf{9.96e3 \pm 8.28e3}$ & $7.18e3 \pm 8.90e3$          & $1.45e3 \pm 1.53e3$          & $4.86e2 \pm 9.98e1$          \\
    \hline
     \multirow{3}{*}{80\%} & AE         & $8.31e3 \pm 9.62e3$ & $\mathbf{1.32e4 \pm 1.20e4}$ & $\mathbf{4.74e3 \pm 5.51e3}$ & $\mathbf{9.11e2 \pm 6.53e2}$ & $5.22e2 \pm 9.65e1$          \\
                           & RBM        & $3.00e4 \pm 2.83e4$ & $1.55e4 \pm 2.14e4$          & $6.26e3 \pm 6.15e3$          & $4.72e3 \pm 8.42e3$          & $5.13e2 \pm 1.41e2$          \\
                           & Ours       & $1.13e4 \pm 1.47e4$ & $1.54e4 \pm 1.46e4$          & $6.48e3 \pm 4.50e3$          & $3.62e3 \pm 2.74e3$          & $\mathbf{3.98e2 \pm 7.89e1}$ \\
    \hline
     \multirow{3}{*}{90\%} & AE         & $1.01e4 \pm 5.01e3$ & $1.25e4 \pm 1.48e4$          & $\mathbf{2.99e3 \pm 2.53e3}$ & $\mathbf{9.04e2 \pm 3.92e2}$ & $5.06e2 \pm 1.74e2$          \\
                           & RBM        & $1.62e4 \pm 1.09e4$ & $1.37e4 \pm 1.05e4$          & $1.22e4 \pm 1.37e4$          & $1.27e3 \pm 7.67e2$          & $4.95e2 \pm 6.30e1$          \\
                           & Ours       & $1.50e4 \pm 1.18e4$ & $\mathbf{1.06e4 \pm 1.03e4}$ & $7.26e3 \pm 8.44e3$          & $3.50e3 \pm 6.38e3$          & $4.89e2 \pm 9.89e1$          \\
    \hline
    \hline
        &   & \multicolumn{5}{c}{FD002}   \\
                     &    & 2\%     & 10\%    & 20\%    & 40\%    & 100\%   \\
    \hline \hline
                           & None       & $6.95e4 \pm 7.74e4$          & $2.02e4 \pm 1.70e4$          & $9.21e3 \pm 4.80e3$          & $4.79e3 \pm 1.95e3$          & $1.99e3 \pm 6.11e2$          \\
    \hline
     \multirow{3}{*}{40\%} & AE         & $1.25e5 \pm 1.32e5$          & $\mathbf{1.21e4 \pm 5.90e3}$ & $8.23e3 \pm 4.71e3$          & $5.80e3 \pm 2.51e3$          & $1.86e3 \pm 5.80e2$          \\
                           & RBM        & $7.16e4 \pm 7.74e4$          & $1.35e4 \pm 8.68e3$          & $\mathbf{6.08e3 \pm 2.18e3}$ & $4.33e3 \pm 2.30e3$          & $\mathbf{1.69e3 \pm 2.60e2}$ \\
                           & Ours       & $1.43e5 \pm 8.00e4$          & $1.45e4 \pm 1.00e4$          & $8.20e3 \pm 3.92e3$          & $\mathbf{4.12e3 \pm 2.29e3}$ & $2.71e3 \pm 1.74e3$          \\
    \hline
     \multirow{3}{*}{60\%} & AE         & $1.01e5 \pm 1.03e5$          & $\mathbf{1.20e4 \pm 5.64e3}$ & $\mathbf{6.71e3 \pm 2.70e3}$ & $5.46e3 \pm 2.16e3$          & $2.08e3 \pm 1.13e3$          \\
                           & RBM        & $3.92e4 \pm 4.56e4$          & $1.29e4 \pm 7.54e3$          & $8.26e3 \pm 4.54e3$          & $\mathbf{3.10e3 \pm 8.06e2}$ & $\mathbf{1.82e3 \pm 3.69e2}$ \\
                           & Ours       & $\mathbf{2.99e4 \pm 2.72e4}$ & $1.25e4 \pm 1.04e4$          & $7.69e3 \pm 6.09e3$          & $3.71e3 \pm 9.67e2$          & $2.78e3 \pm 1.72e3$          \\
    \hline
     \multirow{3}{*}{70\%} & AE         & $7.06e4 \pm 9.35e4$          & $1.92e4 \pm 1.71e4$          & $1.02e4 \pm 9.86e3$          & $5.08e3 \pm 1.72e3$          & $2.59e3 \pm 1.93e3$          \\
                           & RBM        & $6.14e4 \pm 6.51e4$          & $\mathbf{1.30e4 \pm 4.50e3}$ & $7.90e3 \pm 5.90e3$          & $3.57e3 \pm 1.23e3$          & $1.76e3 \pm 2.73e2$          \\
                           & Ours       & $\mathbf{3.33e4 \pm 4.39e4}$ & $1.78e4 \pm 2.29e4$          & $\mathbf{4.74e3 \pm 4.02e3}$ & $\mathbf{2.76e3 \pm 1.38e3}$ & $\mathbf{1.67e3 \pm 1.97e2}$ \\
    \hline
     \multirow{3}{*}{80\%} & AE         & $9.93e4 \pm 1.01e5$          & $1.56e4 \pm 1.63e4$          & $7.76e3 \pm 4.89e3$          & $4.58e3 \pm 2.32e3$          & $1.74e3 \pm 2.47e2$          \\
                           & RBM        & $6.16e4 \pm 6.11e4$          & $\mathbf{1.02e4 \pm 3.96e3}$ & $8.27e3 \pm 3.91e3$          & $\mathbf{3.32e3 \pm 1.34e3}$ & $\mathbf{1.65e3 \pm 2.47e2}$ \\
                           & Ours       & $\mathbf{3.55e4 \pm 5.25e4}$ & $1.04e4 \pm 4.79e3$          & $\mathbf{6.25e3 \pm 4.44e3}$ & $3.88e3 \pm 2.41e3$          & $2.13e3 \pm 4.91e2$          \\
    \hline
     \multirow{3}{*}{90\%} & AE         & $7.85e4 \pm 7.03e4$          & $1.48e4 \pm 6.75e3$          & $8.13e3 \pm 3.31e3$          & $5.43e3 \pm 3.63e3$          & $1.88e3 \pm 3.15e2$          \\
                           & RBM        & $4.95e4 \pm 5.63e4$          & $1.03e4 \pm 6.69e3$          & $7.04e3 \pm 5.64e3$          & $3.44e3 \pm 1.30e3$          & $1.68e3 \pm 2.25e2$          \\
                           & Ours       & $\mathbf{3.75e4 \pm 5.15e4}$ & $\mathbf{8.11e3 \pm 4.40e3}$ & $\mathbf{3.91e3 \pm 2.10e3}$ & $\mathbf{2.41e3 \pm 6.83e2}$ & $\mathbf{1.66e3 \pm 2.01e2}$ \\
    \hline
    \end{tabular}
    \end{adjustbox}
    \caption{\textbf{\ac{rul}-Score results for FD001 and FD002}: We report the mean and standard deviation. The rows represent the \emph{grade of degradation} and the columns the \emph{percent of labeled data} of the data scenario. The second column contains the pre-training method, where \emph{None} is the baseline without any pre-training, and \emph{Ours} is the self-supervised pre-training. The bold results mark the best mean performance for each data scenario. If no result is bold, no approach was able to beat the baseline. Please note that the standard deviation of these results can be misleading as \ac{rul}-Score is an exponential measure.}
    \label{tab:score_12}
    \end{table}
    
    \begin{table}
    \begin{adjustbox}{width=\textwidth,center}
            \centering
            \begin{tabular}{cl|ccccc}
    \hline
        &   & \multicolumn{5}{c}{FD003}   \\
                     &    & 2\%     & 10\%    & 20\%    & 40\%    & 100\%   \\
    \hline \hline
                           & None       & $1.18e8 \pm 2.62e8$          & $8.96e3 \pm 1.20e4$          & $1.31e3 \pm 6.83e2$          & $7.03e2 \pm 3.82e2$          & $7.48e2 \pm 1.71e2$          \\
    \hline
     \multirow{3}{*}{40\%} & AE         & $2.93e4 \pm 3.11e4$          & $2.02e4 \pm 5.66e4$          & $1.40e3 \pm 9.24e2$          & $6.94e2 \pm 3.24e2$          & $7.13e2 \pm 2.08e2$          \\
                           & RBM        & $9.06e4 \pm 1.13e5$          & $\mathbf{4.18e3 \pm 3.15e3}$ & $3.43e3 \pm 7.44e3$          & $9.74e2 \pm 5.84e2$          & $9.88e2 \pm 3.10e2$          \\
                           & Ours       & $\mathbf{1.51e4 \pm 1.31e4}$ & $7.16e3 \pm 9.81e3$          & $\mathbf{1.02e3 \pm 5.83e2}$ & $\mathbf{6.20e2 \pm 2.76e2}$ & $\mathbf{4.95e2 \pm 1.12e2}$ \\
    \hline
     \multirow{3}{*}{60\%} & AE         & $4.03e4 \pm 6.81e4$          & $\mathbf{7.16e3 \pm 1.35e4}$ & $1.39e3 \pm 7.65e2$          & $9.76e2 \pm 3.91e2$          & $7.10e2 \pm 1.44e2$          \\
                           & RBM        & $4.07e4 \pm 6.87e4$          & $9.77e3 \pm 2.35e4$          & $2.64e3 \pm 4.86e3$          & $1.03e3 \pm 6.20e2$          & $9.02e2 \pm 3.54e2$          \\
                           & Ours       & $\mathbf{3.35e4 \pm 4.47e4}$ & $7.24e3 \pm 7.05e3$          & $1.71e3 \pm 1.55e3$          & $\mathbf{6.13e2 \pm 2.42e2}$ & $\mathbf{4.74e2 \pm 1.71e2}$ \\
    \hline
     \multirow{3}{*}{70\%} & AE         & $3.91e4 \pm 3.60e4$          & $7.12e3 \pm 1.50e4$          & $\mathbf{1.05e3 \pm 4.12e2}$ & $1.04e3 \pm 6.46e2$          & $6.78e2 \pm 1.75e2$          \\
                           & RBM        & $6.09e4 \pm 7.69e4$          & $\mathbf{3.49e3 \pm 3.45e3}$ & $1.42e3 \pm 6.43e2$          & $1.01e3 \pm 3.75e2$          & $9.28e2 \pm 2.79e2$          \\
                           & Ours       & $\mathbf{1.65e4 \pm 1.88e4}$ & $1.20e4 \pm 3.14e4$          & $1.16e3 \pm 5.47e2$          & $7.17e2 \pm 2.83e2$          & $\mathbf{5.25e2 \pm 1.27e2}$ \\
    \hline
     \multirow{3}{*}{80\%} & AE         & $4.36e4 \pm 3.58e4$          & $\mathbf{4.11e3 \pm 5.66e3}$ & $1.21e3 \pm 6.58e2$          & $8.46e2 \pm 3.09e2$          & $6.14e2 \pm 1.39e2$          \\
                           & RBM        & $5.96e4 \pm 7.59e4$          & $8.60e3 \pm 1.82e4$          & $1.47e3 \pm 7.13e2$          & $9.10e2 \pm 2.92e2$          & $1.01e3 \pm 2.76e2$          \\
                           & Ours       & $\mathbf{4.20e4 \pm 7.41e4}$ & $6.77e3 \pm 9.98e3$          & $\mathbf{9.61e2 \pm 3.88e2}$ & $\mathbf{5.96e2 \pm 2.08e2}$ & $\mathbf{4.59e2 \pm 1.39e2}$ \\
    \hline
     \multirow{3}{*}{90\%} & AE         & $3.22e4 \pm 4.59e4$          & $\mathbf{3.40e3 \pm 5.39e3}$ & $1.73e3 \pm 1.23e3$          & $8.26e2 \pm 4.42e2$          & $6.66e2 \pm 8.99e1$          \\
                           & RBM        & $8.56e4 \pm 8.92e4$          & $5.24e3 \pm 6.90e3$          & $1.97e3 \pm 1.67e3$          & $9.11e2 \pm 5.46e2$          & $9.92e2 \pm 4.93e2$          \\
                           & Ours       & $\mathbf{1.31e4 \pm 7.50e3}$ & $3.44e3 \pm 4.80e3$          & $\mathbf{1.14e3 \pm 5.58e2}$ & $\mathbf{6.69e2 \pm 2.35e2}$ & $\mathbf{5.55e2 \pm 9.86e1}$ \\
    \hline
    \hline
        &   & \multicolumn{5}{c}{FD004}   \\
                     &    & 2\%     & 10\%    & 20\%    & 40\%    & 100\%   \\
    \hline \hline
                           & None       & $1.63e5 \pm 1.58e5$          & $2.52e4 \pm 2.33e4$          & $1.69e4 \pm 9.37e3$          & $9.15e3 \pm 3.29e3$          & $6.21e3 \pm 1.62e3$          \\
    \hline
     \multirow{3}{*}{40\%} & AE         & $1.44e6 \pm 9.82e5$          & $1.56e6 \pm 8.75e5$          & $2.08e6 \pm 6.41e5$          & $2.27e6 \pm 1.34e6$          & $1.77e6 \pm 9.15e5$          \\
                           & RBM        & $1.60e5 \pm 1.52e5$          & $3.90e4 \pm 3.43e4$          & $1.69e4 \pm 8.72e3$          & $\mathbf{7.79e3 \pm 2.62e3}$ & $6.03e3 \pm 2.25e3$          \\
                           & Ours       & $\mathbf{1.02e5 \pm 4.04e4}$ & $5.25e4 \pm 7.18e4$          & $\mathbf{1.65e4 \pm 1.28e4}$ & $1.12e4 \pm 7.05e3$          & $\mathbf{5.26e3 \pm 1.71e3}$ \\
    \hline
     \multirow{3}{*}{60\%} & AE         & $1.10e6 \pm 8.70e5$          & $1.82e6 \pm 8.91e5$          & $1.71e6 \pm 1.29e6$          & $1.94e6 \pm 1.00e6$          & $1.42e6 \pm 6.38e5$          \\
                           & RBM        & $1.41e5 \pm 1.12e5$          & $3.66e4 \pm 3.26e4$          & $\mathbf{1.29e4 \pm 5.23e3}$ & $8.53e3 \pm 5.18e3$          & $5.76e3 \pm 1.30e3$          \\
                           & Ours       & $\mathbf{3.69e4 \pm 1.60e4}$ & $3.04e4 \pm 1.76e4$          & $1.66e4 \pm 1.35e4$          & $\mathbf{6.67e3 \pm 3.77e3}$ & $\mathbf{4.29e3 \pm 9.92e2}$ \\
    \hline
     \multirow{3}{*}{70\%} & AE         & $1.80e5 \pm 2.11e5$          & $3.86e4 \pm 3.61e4$          & $1.64e4 \pm 7.75e3$          & $1.08e4 \pm 8.82e3$          & $5.73e3 \pm 1.43e3$          \\
                           & RBM        & $1.43e5 \pm 1.39e5$          & $3.68e4 \pm 5.13e4$          & $1.53e4 \pm 9.11e3$          & $9.53e3 \pm 4.82e3$          & $5.38e3 \pm 1.94e3$          \\
                           & Ours       & $\mathbf{3.52e4 \pm 1.87e4}$ & $\mathbf{1.54e4 \pm 6.76e3}$ & $\mathbf{8.43e3 \pm 4.92e3}$ & $\mathbf{8.52e3 \pm 5.06e3}$ & $\mathbf{4.25e3 \pm 1.01e3}$ \\
    \hline
     \multirow{3}{*}{80\%} & AE         & $1.46e6 \pm 1.01e6$          & $1.76e6 \pm 1.09e6$          & $1.37e6 \pm 6.03e5$          & $1.75e6 \pm 6.24e5$          & $1.71e6 \pm 8.83e5$          \\
                           & RBM        & $2.11e5 \pm 2.24e5$          & $3.41e4 \pm 2.14e4$          & $1.42e4 \pm 8.29e3$          & $9.48e3 \pm 5.50e3$          & $5.13e3 \pm 1.09e3$          \\
                           & Ours       & $\mathbf{2.18e4 \pm 2.11e4}$ & $\mathbf{1.61e4 \pm 1.51e4}$ & $\mathbf{1.36e4 \pm 1.10e4}$ & $\mathbf{5.87e3 \pm 2.13e3}$ & $\mathbf{4.25e3 \pm 1.01e3}$ \\
    \hline
     \multirow{3}{*}{90\%} & AE         & $1.58e5 \pm 1.73e5$          & $3.28e4 \pm 2.23e4$          & $1.59e4 \pm 1.02e4$          & $1.04e4 \pm 5.96e3$          & $5.74e3 \pm 4.26e2$          \\
                           & RBM        & $1.46e5 \pm 1.24e5$          & $2.51e4 \pm 2.21e4$          & $1.61e4 \pm 1.16e4$          & $1.09e4 \pm 5.88e3$          & $5.60e3 \pm 1.27e3$          \\
                           & Ours       & $\mathbf{5.05e4 \pm 7.02e4}$ & $\mathbf{1.14e4 \pm 7.05e3}$ & $\mathbf{1.17e4 \pm 1.27e4}$ & $\mathbf{5.52e3 \pm 2.95e3}$ & $\mathbf{4.04e3 \pm 8.03e2}$ \\
    \hline
    \end{tabular}
    \end{adjustbox}
    \caption{\textbf{\ac{rul}-Score results for FD003 and FD004}: Please consult table \ref{tab:score_12} for further information.}
    \label{tab:score_34}
    \end{table}
    
    \subsection{Discussion of Findings}
    \label{sec:R:Discussion}
        Our results revealed several points worthy of further discussion.
        First of all, there is the matter of minimal baseline performance drops when using as few as 40\% of the labeled data.
        A possible explanation for this is the fact that the \ac{cmapss} dataset is the product of a simulation which may lead to little variation between the individual time series.
        If the variation between time series is small, the network needs less data to learn useful patterns from it.
        The differences between the subsets may be explained by the varying number of available time series per subset.
        Additional experiments where the absolute number of labeled data is varied instead of a percentage can be used to confirm this hypothesis.
        
        The next point would be the discrepancies between \ac{rmse} and \ac{rul}-Score performance best seen on FD004 for the \ac{ae} pre-training.
        Even though the \ac{rmse} performance is similar to the other approaches, the \ac{rul}-Score is much higher.
        This phenomenon can be explained through the asymmetric nature of the \ac{rul}-score, i.e. that late predictions incur a higher penalty than early ones.
        The \ac{ae} pre-trained models seem to make predominantly late predictions, even though the absolute difference from the real \ac{rul} is similar to the other approaches.
        
        Another interesting finding is, that \ac{rbm} pre-training is competitive with \ac{ae} pre-training, even though it pre-trains only the first layer.
        Both methods use a reconstruction-based pre-training task, which makes the comparison even more interesting.
        An explaining factor could be that the bottleneck size of the autoencoder was not optimized as a hyperparameter, although it has a significant influence on the reconstruction capabilities of the \ac{ae}.
        We did not optimize it because changing the bottleneck size would change the number of model parameters making the comparison with other methods difficult.
        A bigger bottleneck may increase performance because of the increased parameter count independently from pre-training.
        Additional experiments where the bottleneck size is optimized for the autoencoder and then compared to the other approaches on the same model architecture could prove this hypothesis.
        
        The last point is the downward trend in performance for self-supervised pre-training with respect to the grade of degradation.
        At least on FD002 and FD004, we can observe that performance drops at lower grades of degradation and even makes our approach worse than the baseline in some cases.
        This problem may lie in the nature of the pre-text task we choose.
        It is based on the assumption that differences in the features of two time frames are correlated with the difference in \ac{rul}.
        If we take the piece-wise linear degradation model of \ac{cmapss} at face value, there are no differences in \ac{rul} above $\textsc{RUL}_{max}$.
        Consequently, there should not be a difference in the features either.
        Our pre-text task on the other hand is only accurate for a linear degradation model as we cannot take $\textsc{RUL}_{max}$ into account on unlabeled data.
        Looking at the percentage of time frames above and below $\textsc{RUL}_{max}$ for different grades of degradation, we can see that in FD004 only 11\% of the training data has a label below a $\textsc{RUL}_{max}$ of 125 at 40\% grade of degradation.
        Coincidentally, the trend is most obvious on this subset.
        The training data of FD001, FD002, and FD003 has 33\%, 23\%, and 16\% labels below 125 at 40\% grade of degradation.
        Less data with labels below $\textsc{RUL}_{max}$ makes the approximation of our pre-text task less accurate.

\section{Conclusion \& Future Work}
\label{sec:Conclusion}
    We presented a study of three \ac{ssl} approaches on the NASA \ac{cmapss} dataset under improved, more realistic evaluation conditions.
    These conditions take into account that realistic, unlabeled data does not contain features near the point of failure.
    Concerning our first research question, our results show that, contrary to previous studies, \ac{ssl} does not always improve the accuracy of \ac{rul} estimation.
    This underlines the importance of a representative validation set to identify negative outcomes which may not always be available in settings with few labeled time series.
    
    In answering our second research question, we have shown that our \ac{ssl} approach, based on self-supervised pre-training, has superior performance compared to the competing approaches under certain conditions.
    More work is necessary to replicate these findings on other datasets.
    Nevertheless, our approach was not able to beat the baseline performance reliably under all data scenarios.
    Most notably, the performance dropped when the grade of degradation was low.
    
    Future work includes conducting the experiments outlined in the discussion section to test the proposed explanations to the observed phenomena.
    Additionally, advanced techniques from the field of metric learning, e.g. hard sample mining or triplet loss, could be used to improve the self-supervised pre-training.
    Theoretically, our approach could be used for Unsupervised \ac{da}, too, as it shares many characteristics with \ac{ssl}.
    Unsupervised \ac{da} is of high interest for \ac{rul} estimation as labeled and unlabeled data often do not share the same domain.
    Investigating the effectiveness of our approach for general, non-linear degradation processes, e.g. tool wear estimation, is another direction for future work.
    
\section*{Acknowledgement}

    This work was funded by the Technische Universität Berlin with support of IAV GmbH.

\bibliography{paper}

\end{document}